\pdfoutput=1

\documentclass[11pt]{article}

\usepackage{EMNLP2022}

\usepackage{times}
\usepackage{latexsym}

\usepackage[T1]{fontenc}

\usepackage[utf8]{inputenc}

\usepackage{microtype}

\usepackage{inconsolata}

\usepackage{adjustbox}
\usepackage{amsfonts}
\usepackage{amsmath}
\usepackage{amssymb}
\usepackage{bm}
\usepackage{booktabs}
\usepackage{colortbl}
\usepackage{comment}
\usepackage{dsfont}
\usepackage{enumitem}
\usepackage{graphicx} 
\usepackage{hyperref}
\usepackage{mathtools}
\usepackage{microtype}
\usepackage{multicol}
\usepackage{multirow}
\usepackage{nicefrac}
\usepackage{lscape}
\usepackage{scalerel}
\usepackage{subcaption}
\usepackage{tabularx}
\usepackage[most]{tcolorbox}
\usepackage{xcolor}
\usepackage{xspace}

\newcommand\blankfootnote[1]{%
  \let\thefootnote\relax\footnotetext{#1}%
  \let\thefootnote\svthefootnote%
}
\newcommand\myfontsize{\fontsize{9.6pt}{10.5pt}\selectfont}
\newcommand{\aclcr}[1]{#1}

\newcommand{\ssc}[1]{{\small \sc #1}\xspace}
\newcommand{\bssc}[1]{{\small \sc \textbf{#1}}\xspace}
\newcommand{\fssc}[1]{{\scriptsize \sc #1}\xspace}
\newcommand{\prompt}{{\ssc{Prompt}}\xspace}
\newcommand{\model}{{\ssc{Model}}\xspace}
\newcommand{\prompttuning}{{\ssc{PromptTuning}}\xspace}
\newcommand{\bprompttuning}{{\bssc{PromptTuning}}\xspace}
\newcommand{\modeltuning}{{\ssc{ModelTuning}}\xspace}
\newcommand{\bmodeltuning}{{\bssc{ModelTuning}}\xspace}

\newcommand{\smallsup}[1]{\scaleto{\text{#1}}{4pt}}
\newcommand{\mtfive}{{mT5}\xspace}
\newcommand{\mcfour}{{mC4}\xspace}
\newcommand{\sprouge}{{\ssc{SP-RG}}\xspace}

\newcommand{\bsprouge}{{\bssc{SP-RG}}\xspace}
\newcommand{\fsprouge}{{\fssc{SP-RG}}\xspace}

\newcommand{\wikilinguazero}{{\mbox{\ssc{WikiLingua-0}}}\xspace}

\definecolor{myblue}{HTML}{2B79B0}
\definecolor{myorange}{HTML}{FB7F36}
\definecolor{mygreen}{HTML}{389E3B}

\title{Overcoming Catastrophic Forgetting in \\ Zero-Shot Cross-Lingual Generation}

\author{Tu Vu$^{1,2}$$^\bigstar$, Aditya Barua$^{1}$, Brian Lester$^{1}$, Daniel Cer$^{1}$, Mohit Iyyer$^{2}$, Noah Constant$^{1}$\\
  Google Research$^1$\\
  University of Massachusetts Amherst$^2$ \\
  \texttt{\{ttvu,\,adityabarua,\,brianlester,\,cer,\,nconstant\}@google.com}\\ 
  \texttt{\{tuvu,\,miyyer\}@cs.umass.edu}\\}
  
\begin{document}
\maketitle

\begin{abstract}
\label{section:abstract}
In this paper, we explore the challenging problem of performing a generative task in a target language when labeled data is only available in English, using summarization as a case study. We assume a strict setting with no access to parallel data or machine translation and find that common transfer learning approaches struggle in this setting, as a generative multilingual model fine-tuned purely on English catastrophically forgets how to generate non-English. Given the recent rise of parameter-efficient adaptation techniques, we conduct the first investigation into how one such method, prompt tuning~\cite{BLester21}, can overcome catastrophic forgetting to enable zero-shot cross-lingual generation. Our experiments show that parameter-efficient prompt tuning provides gains over standard fine-tuning when transferring between less-related languages, e.g., from English to Thai. However, a significant gap still remains between these methods and fully-supervised baselines. To improve cross-lingual transfer further, we explore several approaches, including: %
(1) mixing in unlabeled multilingual data, and (2) explicitly factoring prompts into recombinable language and task components. Our approaches can provide further quality gains, suggesting that robust zero-shot cross-lingual generation is within reach.
\blankfootnote{$\bigstar$ Work done as a student researcher at Google Brain.}
\end{abstract}
\section{Introduction}
\label{section:introduction}
\begin{figure}[t!]
\centering
\includegraphics[width=0.48\textwidth]{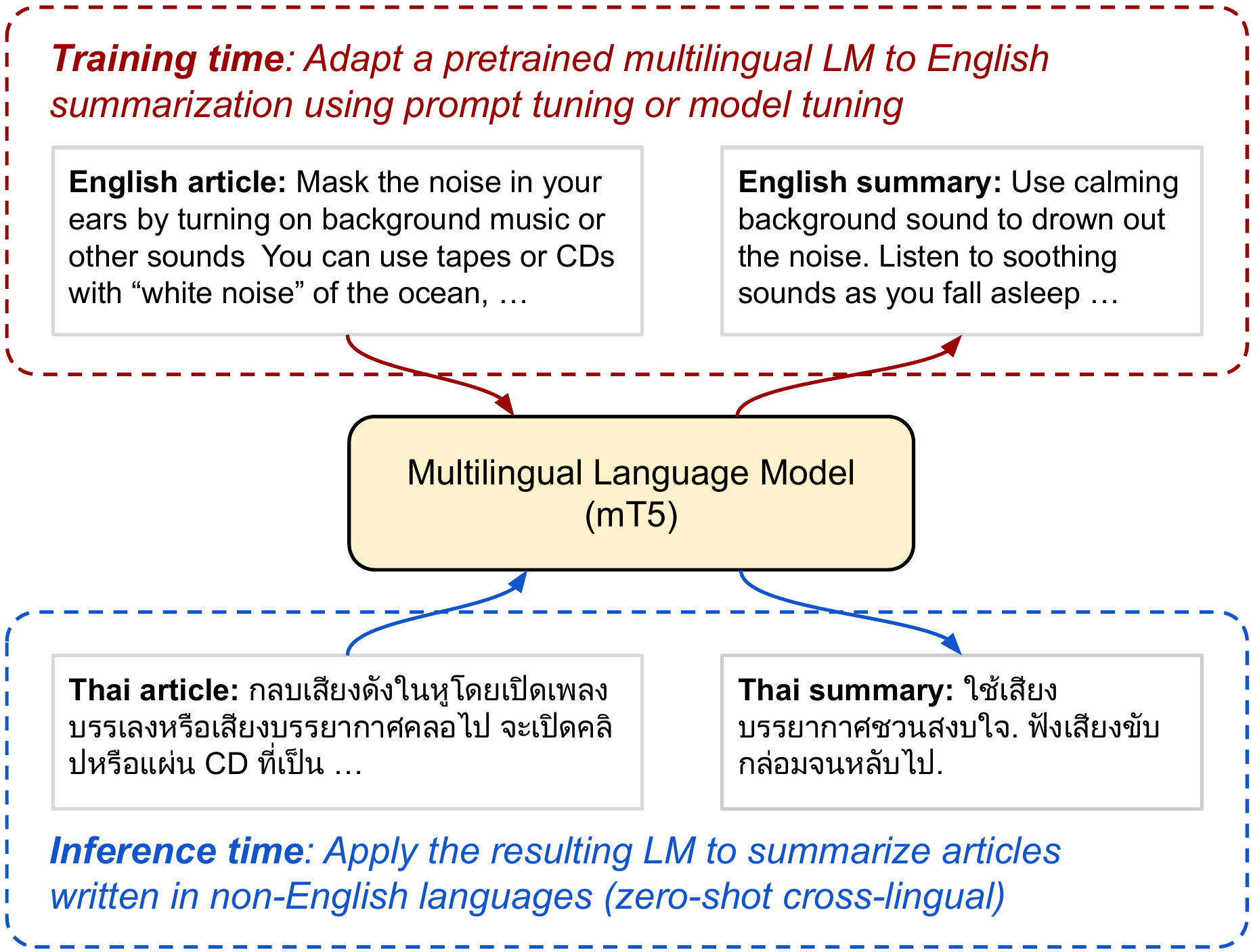}
\caption{A demonstration of \ssc{WikiLingua-0}, a challenging zero-shot cross-lingual generation (\ssc{XGen}) task,  which requires a model to learn a generative task from labeled data in one language (i.e., English), and then perform the equivalent task in another language at inference time.}
\label{figure:wikilingua0}
\end{figure}
Cross-lingual language understanding is an important area of ongoing research \cite{AConneau20,JHu20,SRuder21}. With vastly differing amounts of data (both labeled and unlabeled) available across languages, there is significant value to developing techniques that can transfer knowledge from higher-resource languages to improve performance in lower-resource languages. \emph{Zero-shot} cross-lingual benchmarks push on the limiting case where no labeled data is available in the target language. Remarkable progress has been made on zero-shot cross-lingual tasks by scaling up the size of pre-trained multilingual models \cite{AConneau20,LXue21}. However, prior work has focused nearly exclusively on \emph{non-generative tasks} (e.g., classification, extractive question answering, and sequence labeling).

In this paper, we turn our attention to zero-shot cross-lingual \emph{generation}, or ``\ssc{XGen}'', which requires a model to learn a generative task from labeled data in one language (typically English), and then perform the equivalent generative task in another language. This problem is particularly challenging because generative models trained on one language are known to exhibit catastrophic forgetting, losing the ability to generate coherent text in other languages \cite{LXue21,KMaurya21,SShakeri21}. In particular, we focus on the relatively under-explored task of zero-shot cross-lingual summarization. %
We construct a new zero-shot task \mbox{\ssc{WikiLingua-0}} from the \ssc{WikiLingua} dataset~\cite{FLadhak20}, allowing us to test \ssc{XGen} capabilities across 18 languages. We motivate a new evaluation metric for our task, \ssc{SP-Rouge}, and show that it correlates well with human judgments of summary quality.

\citet{KMaurya21} show improved performance on \ssc{XGen} tasks by freezing model parameters in the input and output layers during fine-tuning. Inspired by recent parameter-efficient adaptation techniques \cite{NHoulsby19,EZaken21,XLi21,BLester21}, we take this approach further: can we overcome catastrophic forgetting by freezing \emph{all} of the pre-trained model parameters, and only tuning a much smaller set of task-specific parameters? Parameter-efficient tuning methods are particularly appealing for multilingual NLP, as they would enable reuse of a single frozen model across many combinations of task and language, reducing storage and serving costs.%

To this end, we conduct %
the first investigation of the \ssc{XGen} performance of \prompttuning~\cite{BLester21}, a simple parameter-efficient adaptation technique that limits learned parameters to a set of virtual tokens prepended to the text input. We compare \prompttuning with standard fine-tuning (or \modeltuning, where all model weights are tuned) across different languages and model scales. %
We find that increasing model size and decreasing tunable parameter capacity are key for overcoming catastrophic forgetting. Despite its inferior performance on the training language (English), \prompttuning with scale typically outperforms \modeltuning when evaluated on non-English languages, especially on languages more distantly related to English, such as Thai. This corroborates previous findings~\cite{XLi21,BLester21} that parameter-efficient methods are more robust to domain shifts between training and inference.

Motivated by our initial findings, we investigate two approaches to further improve the \ssc{XGen} performance of \prompttuning and \modeltuning. %
Our first approach involves mixing unlabeled data in the target language into the supervised training stage. We show this dramatically alleviates catastrophic forgetting on \wikilinguazero. %
We also introduce a novel approach, ``factorized prompts'', %
which is specifically designed for \prompttuning. We train prompts on a multi-task multilingual mixture, where each prompt is factorized into composable language and task modules---the first half of the prompt encodes language knowledge, while the second half captures language-agnostic task knowledge. During inference in the zero-shot cross-lingual setting, the source language module is replaced with the target language module, while the task module remains unchanged. We demonstrate that factorized prompts provide an effective means of improving \ssc{XGen} performance. %

To summarize, our main contributions are: 
\begin{itemize}
    \item We present the first large-scale empirical investigation of parameter-efficient \prompttuning and standard \modeltuning for zero-shot cross-lingual generation (\ssc{XGen}). We show that increasing model scale and decreasing tunable parameter capacity are key for overcoming catastrophic forgetting on \ssc{XGen}.
    \item We propose \wikilinguazero, a challenging \ssc{XGen} benchmark and an associated \mbox{\ssc{SP-Rouge}} evaluation metric, which we hope will facilitate future work evaluating multilingual summarization.
    \item We show that mixing in unsupervised multilingual data can boost \ssc{XGen} performance, and are the first to combine this approach with \prompttuning.
    \item We propose ``factorized prompts'', a novel approach that can also help \prompttuning overcome severe catastrophic forgetting.
    \item To facilitate future work, we  release our data, pretrained models, and code at:\\
    \href{https://github.com/google-research/prompt-tuning/tree/main/prompt_tuning/x_gen}{\ttfamily\myfontsize\aclcr{https://github.com/google-research/ prompt-tuning/tree/main/prompt\_tuning/\\x\_gen}}.
\end{itemize}

\section{Challenge of zero-shot cross-lingual generation}
\label{section:challenge_wikilingua0}

Much recent progress in multilingual NLP has been driven by zero-shot cross-lingual benchmarks that require a model to perform classification~\cite{AConneau18,YYang19}, extractive QA~\cite{MArtetxe20,PLewis20,JClark20}, or sequence labeling~\cite{XPan17}.\footnote{We refer the interested reader to Appendix~\ref{appendix:xlingeval} for a comprehensive comparison of \fssc{ModelTuning} and \fssc{PromptTuning} on these benchmarks. Overall, we find that \fssc{ModelTuning} typically performs better than \fssc{PromptTuning}, although \fssc{PromptTuning} at scale matches the performance of \fssc{ModelTuning} on English and can yield better results on some languages.} Here, we are interested in a more challenging task of zero-shot cross-lingual generation (\ssc{XGen}) where a model is trained on a generative task in one language (typically English), and then asked to perform the equivalent task in another language during inference. We construct a novel zero-shot cross-lingual summarization task and show that state-of-the-art text-to-text models adapted using \modeltuning and \prompttuning techniques are not able to successfully perform our task. Our analysis reveals that both techniques suffer from catastrophic forgetting, causing them to often generate text in the wrong language. %

\subsection{Problem formulation}
\label{subsection:problem_formultation}

\paragraph{Defining \bssc{WikiLingua-0} zero-shot cross-lingual summarization:} %
We leverage the \ssc{WikiLingua} dataset~\cite{FLadhak20,SGehrmann21} to create a novel zero-shot cross-lingual summarization task, which we dub \mbox{\ssc{WikiLingua-0}}.\footnote{Note that the original \fssc{WikiLingua} task is not suitable for direct use in our \fssc{XGen} setting, as it aims to generate English summaries from non-English articles.} While \ssc{WikiLingua} provides labeled training data in 18 languages (including English), we are interested in a more realistic experimental setup where no training data is provided in non-English languages, as it is less practical to obtain labeled data for real low-resource languages.\footnote{While one might rely on machine translation (\fssc{MT}) to obtain labeled data in a language of interest, this is not particularly appealing due to: (i) extra computation required, (ii) varied translation quality across languages~\cite{SRuder21}, (iii) potential loss of discourse structure~\cite{JLi14}, and (iv) limited understanding of black box \fssc{MT} systems.}
As such, we discard all training data for non-English languages, with the exception of ablation experiments, and cast \ssc{WikiLingua} as training a model with English summarization data and feeding it non-English articles during zero-shot evaluation.\footnote{See \citet{FLadhak20} for data statistics.}
\paragraph{Defining \bsprouge for multilingual summarization evaluation:} \textsc{rouge}~\cite{CLin04} has been the metric of choice for evaluating summarization systems. However, it assumes that the input text uses spaces to separate words, which is not the case for many languages (e.g., Chinese, Japanese, and Thai).\footnote{In preliminary experiments, we found that standard \fssc{Rouge} yielded extremely poor \fssc{Rouge} scores in many languages, despite systems producing reasonably good summaries.}
One possible solution is to use language-specific tokenizers, as done in~\citet{AConneau19}.
To avoid language-specific preprocessing, we use SentencePiece sub-word tokenization~\cite{TKudo18}, which is data-driven and language independent.\footnote{\citet{NGoyal21} also use a similar approach for \fssc{BLEU}~\cite{KPapineni02}.}
We call our metric \ssc{SP-Rouge} (SentencePiece-based \ssc{Rouge}) or \sprouge for short, and report \mbox{\sprouge-\ssc{Lsum}} in our experiments.\footnote{\fssc{Rouge-Lsum} is the summary-level \fssc{Rouge-L} metric used in~\citet{ASee17}.
}
In Appendix \ref{appendix:correlation}, we demonstrate that \ssc{SP-Rouge} produces a similar correlation to human judgments as \ssc{BLEURT}~\cite{TSellam20} while being significantly more computationally efficient.

\subsection{Experimental setup}
\subsubsection{Baselines}
In addition to vanilla \modeltuning and \prompttuning, we consider the following baselines:
\paragraph{\bssc{Lead-64}:} This baseline simply copies the first $64$ SentencePiece tokens from the input article.\footnote{In our preliminary experiments, $n=64$ performed best among a range of values $\{32, 64, 128, 256\}$.}
\paragraph{\bssc{trans-train}:} We perform \modeltuning or \prompttuning on \wikilinguazero English summarization data that is translated into the target language using \ssc{Google Translate}.%
\paragraph{\bssc{trans-test}:} We train on English summarization data and evaluate on validation data that is translated from the target language to English.
\paragraph{\bssc{sup \& sup-all}:} To ablate the impact of using the labeled training data provided in the original \ssc{WikiLingua} dataset for all languages, we either train on supervised data for each individual target language (\ssc{sup}) or a mixture of supervised data from all languages (\ssc{sup-all}).\footnote{This is an upper bound and is not in the \ssc{XGen} setting.}

\subsubsection{Training and implementation details}
\label{para:hyper-params}
We perform \modeltuning and \prompttuning on top of  pretrained \mtfive checkpoints~\cite{LXue21} of all sizes: \ssc{Small}, \ssc{Base}, \ssc{Large}, \ssc{XL}, \ssc{XXL},\footnote{These are 300M, 580M, 1.2B, 3.7B, and 13B parameters.} using \ssc{T5X} \cite{ARoberts22}. For \prompttuning, we create an \ssc{LM} adapted version of these checkpoints by further training them for 100K steps with the ``prefix LM'' objective~\cite{CRaffel20} using \mcfour~\cite{LXue21} data for all languages.\footnote{A similar approach was used in~\citet{BLester21} for \fssc{PromptTuning} with \fssc{T5}.} Except for ablations, we use $100$ prompt tokens and initialize the prompt by sampling from the vocabulary embeddings. %
Training inputs and targets are clipped to $1024$ and $512$ SentencePiece tokens, respectively. We always train for $100{,}000$ steps for both \modeltuning and \prompttuning. We save a checkpoint every $5{,}000$ steps and report results on the model checkpoint corresponding
to the highest performance on a target language using 250 validation examples for all languages.%
\footnote{For inference, we use beam search with a beam size of $4$ and a length penalty of $\alpha=0.6$. To avoid severe penalties for predictions that repeat a phrase indefinitely, we heuristically remove all but one occurrence of any prediction-final repeated substring.}

\subsection{Results and Discussion}
\begin{figure*}[t!]
\centering
\includegraphics[width=\textwidth]{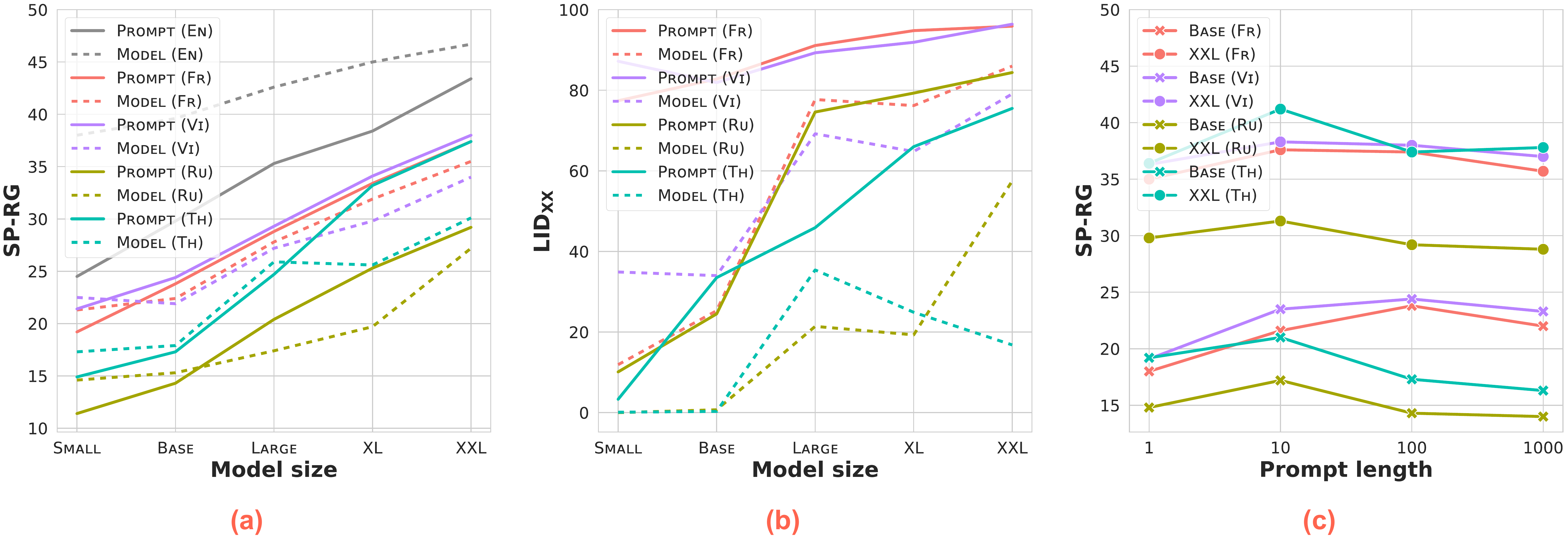}
\caption{\textbf{(a)} Zero-shot \ssc{XGen} summarization quality (\sprouge) and \textbf{(b)} target language accuracy (\ssc{LID$_\mathrm{XX}$}) of \prompttuning and \modeltuning models across five model sizes and four target languages: French (\ssc{Fr}), Vietnamese (\ssc{Vi}), Russian (\ssc{Ru}), and Thai (\ssc{Th}). English (\ssc{En}) performance is provided as a point of comparison, but is no longer a zero-shot task. \textbf{(c)} The effect of prompt length on \prompttuning performance at \ssc{Base} and \ssc{XXL} model sizes.}
\label{figure:model_adaptation_wikilingua0}
\vspace{-2mm}
\end{figure*}

\paragraph{\bssc{WikiLingua-0} is challenging for both \bmodeltuning and \bprompttuning:} Our zero-shot evaluation results on \ssc{WikiLingua-0} for French (\bssc{Fr}), Vietnamese (\bssc{Vi}), Russian (\bssc{Ru}), and Thai (\bssc{Th}) are shown in Figure~\ref{figure:model_adaptation_wikilingua0}a.\footnote{See Table~\ref{tbl:challenge_wikilingua0} in Appendix~\ref{appendix:model_adaptation_wikilingua0} for full results (including variance statistics) and Table~\ref{tbl:wikilingua} in Appendix~\ref{appendix:xlingeval} for results across all target languages.} %
For comparison, we also include results on English. Overall, we find that zero-shot inference on an unseen language leads to a substantial performance drop for both model adaptation techniques, especially when feeding in articles in non-Latin script languages like Russian and Thai. Consistent with the findings in~\citet{SAn22} for other generative tasks, we find that \prompttuning, even with scale, falls far below \modeltuning on monolingual English summarization.\footnote{This is somewhat surprising since across the other tasks we tried above, \fssc{PromptTuning} at \fssc{XXL} can match the performance of \fssc{ModelTuning} when evaluated on English.}

\paragraph{\bprompttuning is better on larger language shifts:} %
Interestingly, \prompttuning is competitive with or out-performs \modeltuning when evaluated on other languages. For instance, at the \ssc{XXL} scale, \prompttuning outperforms \modeltuning by a large margin of $+7.3$ \ssc{SP-Rouge} ($37.4$ vs.~$30.1$) on Thai. A closer look at these results reveals an interesting pattern: as model size increases, \prompttuning usually produces better results than \modeltuning when there is a significant language shift at inference time (e.g., from English to a non-Latin script language).\footnote{With the exception of a few languages (e.g., Chinese).}  This corroborates the view in~\citet{BLester21} that \modeltuning may be over-parameterized and thus more prone to overfit the training task and less robust to domain shifts.

\paragraph{Both \bmodeltuning and \bprompttuning suffer from catastrophic forgetting and this effect is more pronounced for \modeltuning:} When performing zero-shot evaluation on non-English languages, we discover that both \modeltuning and \prompttuning often partially summarize non-English articles into English instead of the target language. This suggests that they suffer from overfitting on the training task. To probe more deeply into this problem, we evaluate performance for each saved checkpoint, and additionally measure: (i) \ssc{LID$_{lang}$}---the average confidence score given by \texttt{cld3}\footnote{\url{https://github.com/google/cld3}} when detecting the language \ssc{$lang$}, and (ii) \ssc{ASCII}---the average percentage of \ssc{ASCII} characters present in the model's predictions, with a higher value indicating a larger amount of English in the model's output for non-Latin script languages.
\begin{figure}[t!]
\centering
\includegraphics[width=0.48\textwidth]{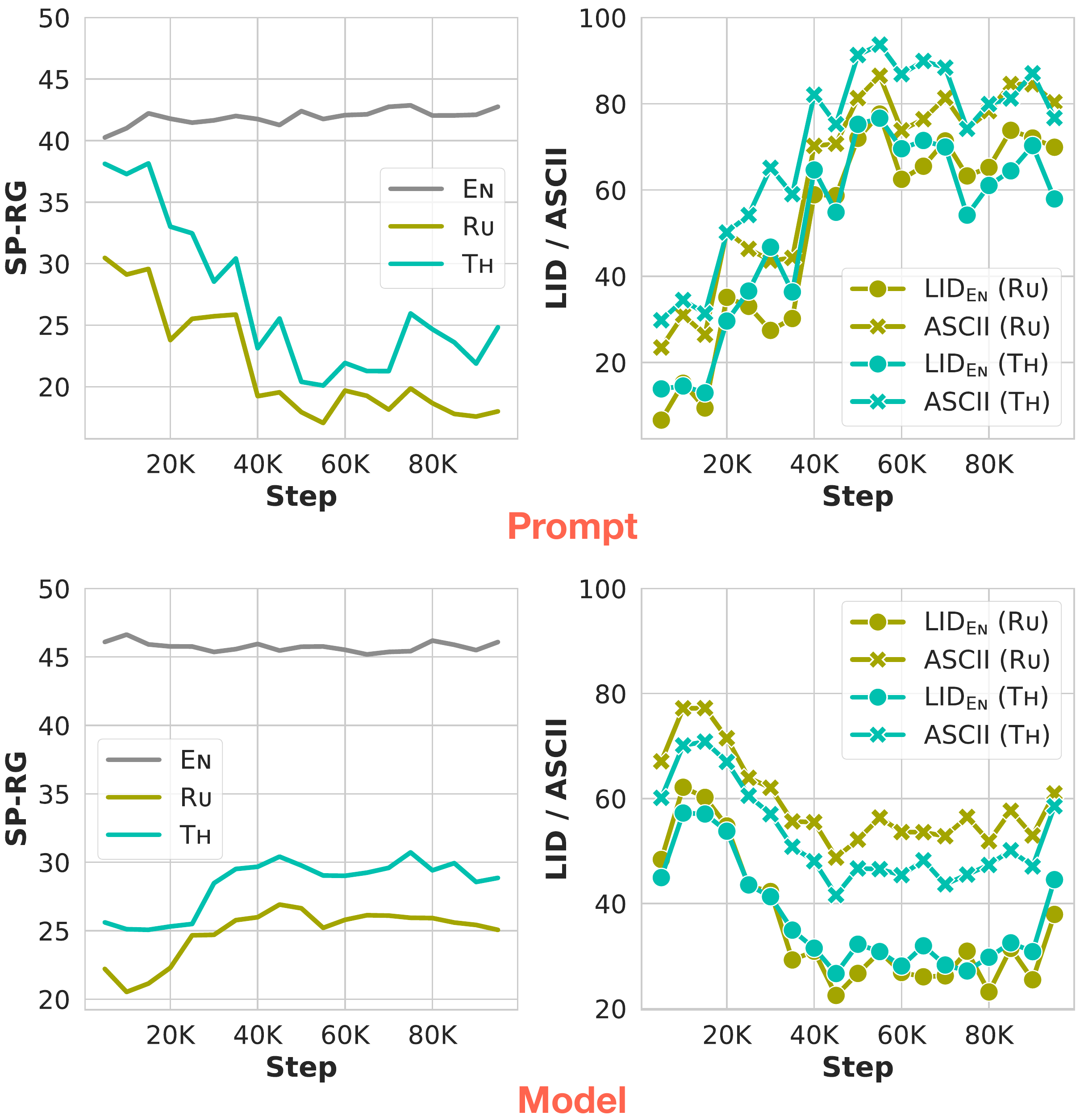}
\caption{Learning curves showing how \prompttuning (top) and \modeltuning (bottom) progress in terms of summarization quality (left) and unwanted English output (right), at the \ssc{XXL} model size. Note, \modeltuning  quality is lower overall, and predictions contain high (>40\%) levels of unwanted \ssc{ASCII}.}
\label{figure:learning_curves}
\end{figure}
Figure \ref{figure:learning_curves} shows our evaluation results as training progresses. For \prompttuning, we observe a clear ``deteriorating'' trend, where the longer the prompt is tuned on English, the more unwanted English is generated, and the lower summarization quality becomes for Russian and Thai. For \modeltuning, even by the first checkpoint, the model has already heavily overfit to English, outputting >$60$\% \ssc{ASCII} for Russian and Thai inputs. There is a modest recovery later in training, but quality as measured by \ssc{SP-Rouge} remains low.

\paragraph{Bigger models are less prone to forget:} In Figure~\ref{figure:model_adaptation_wikilingua0}b, we observe that moving to larger model sizes mitigates catastrophic forgetting to a large extent. This is true both for \modeltuning (in line with the findings of \citet{LXue21}), as well as for \prompttuning. For example, at \ssc{Small} size, \modeltuning and \prompttuning only successfully generate Russian text $0.0$\% and $10.1$\% of the time respectively, whereas at \ssc{XXL} size, these numbers jump to $57.5$\% and $84.4$\%.

\paragraph{Too much capacity is harmful:} Figure~\ref{figure:model_adaptation_wikilingua0}c shows an interesting ``paradox of capacity'' with regard to the prompt length for \prompttuning. On the one hand, greater capacity (in the form of longer prompts) clearly helps to better learn the summarization task. On the other hand, the greater the capacity to learn from English training data, the more the model forgets other languages. We observe that at the beginning of training, the little amount of English introduced in generated outputs is eclipsed by the improvement in summarization quality, which results in a better \ssc{SP-Rouge} score. As training continues, however, the increased capacity becomes harmful as more and more English is introduced in the model's output, which dominates the improvement in summarization quality and leads to lower \ssc{SP-Rouge}. For each language and model size, we observe a critical point past which adding extra capacity becomes harmful. For instance, in Thai at the \ssc{XXL} size, increasing capacity from $1$ to $10$ prompt tokens improves summarization quality (\ssc{SP-Rouge}~$+4.8$) despite a drop in language accuracy (\ssc{LID$_{\text{Th}}$}~$-8.0$), and increasing capacity further to $100$ tokens hurts both metrics.
\begin{figure}[t!]
\centering
\includegraphics[width=0.48\textwidth]{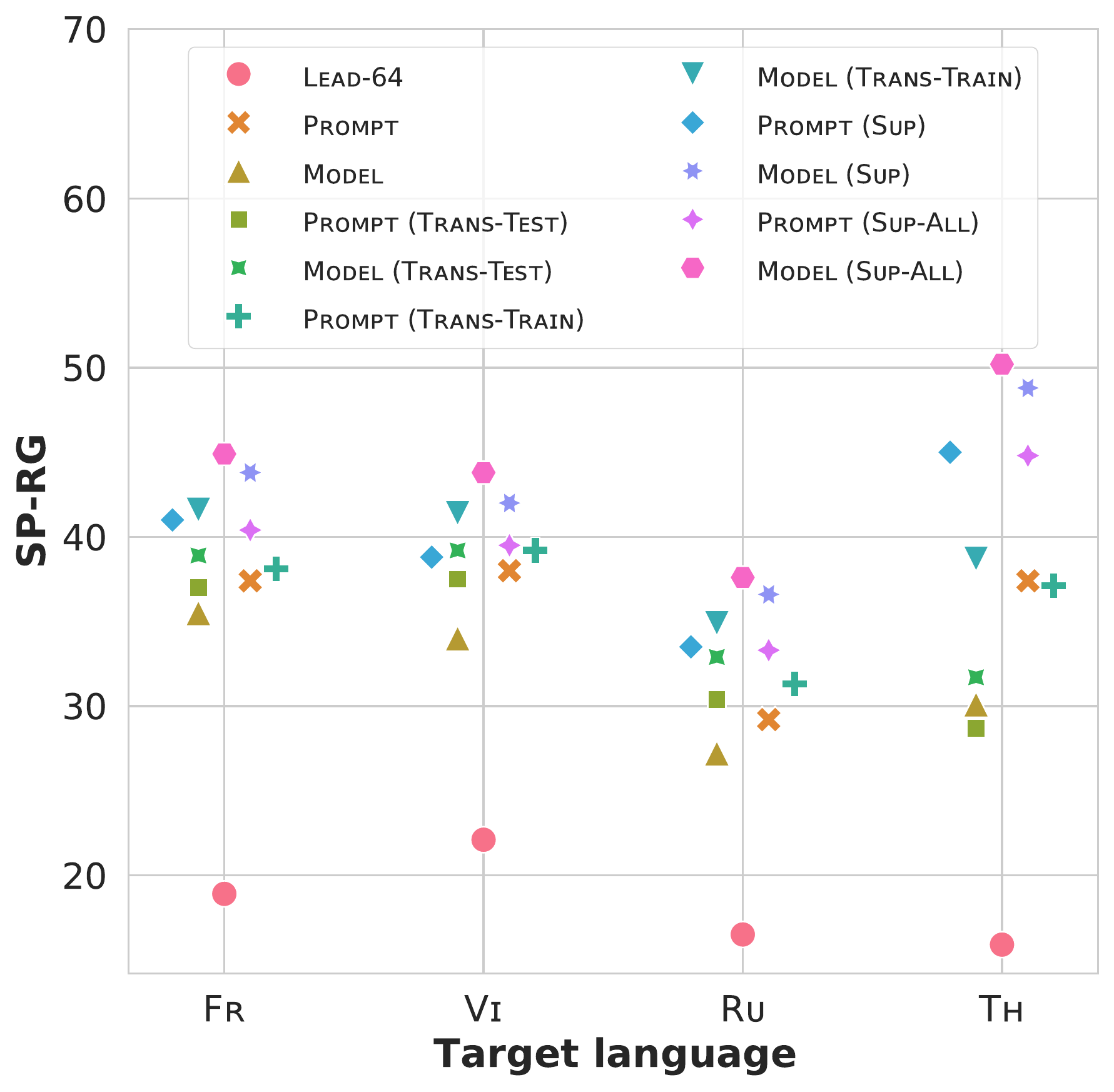}
\caption{\ssc{SP-Rouge} scores of our baselines (\mbox{\ssc{Lead-64}}, \prompttuning, \modeltuning) at the \ssc{XXL} model size, in the zero-shot \ssc{XGen} setting. For comparison, we also show the headroom available if a machine translation system is used (\ssc{trans-train}, \ssc{trans-test}), or if gold data in target languages is used (\ssc{sup}, \ssc{sup-all}).}
\label{figure:baselines}
\vspace{-2mm}
\end{figure}
\paragraph{Significant headroom remains:} The supervised baselines in Figure~\ref{figure:baselines} highlight that significant headroom remains on this \ssc{XGen} task. When tuning the \ssc{XXL} model directly on supervised training data in all languages, \ssc{SP-Rouge} scores are between $+5.8$ (\ssc{Vi}) and $+12.8$ points (\ssc{Th}) higher than our highest zero-shot results. We also note that for some languages, like Thai, the supervised baseline greatly exceeds any approach using machine translation. This highlights that machine translation quality is still low in some languages, so pursuing stronger zero-shot solutions is worthwhile.%

\section{Mitigating catastrophic forgetting}
\label{section:method}
We have seen that increasing model scale and decreasing tunable parameter capacity are both effective in improving \ssc{XGen} performance. Can we obtain further gains by devising methods that explicitly tackle catastrophic forgetting? Here, we investigate two approaches: mixing unlabeled training data with English supervised data, %
and factorizing the learned prompts into composable language and task modules. We show that %
both methods can provide substantially better results when there is severe catastrophic forgetting. %
Below, we describe each method and analyze our findings in detail. 

\subsection{Methods}
\label{section:method_description}
\paragraph{Mixing in unlabeled training data:} This approach involves multi-task learning by mixing an unsupervised training task (\bssc{Unsup}) into the \ssc{WikiLingua-0} data. Mixing is controlled by a mixing rate $\kappa$, resulting in a final mixture that is $\kappa$\% \ssc{Unsup} data and $(100-\kappa)$\% \ssc{WikiLingua-0}. As a data augmentation scheme, this method can be applied in all settings. We use the span corruption pretraining objective from T5~\cite{CRaffel20} with \mcfour data. We create separate multilingual datasets for each target language (\bssc{Mix-Unsup}) as well as a single multilingual dataset that includes all of the \ssc{WikiLingua-0} languages (\bssc{Mix-Unsup-All}). Our goal is to encourage the model not to forget about other languages during training on English summarization. In our experiments, we use $\kappa=1$.\footnote{In our preliminary experiments, $\kappa=1$ performed best among a range of values $\{1, 5, 10, 30, 50\}$. We conjecture that a value of $\kappa>1$ would prevent the model from focusing on the main task of summarization as more unsupervised data is added.} An alternative approach is to perform model or prompt tuning on an intermediate task before tuning on \ssc{Wikilingua-0}. This intermediate tuning approach has been used to boost performance on English tasks for both \modeltuning~\cite{JPhang19,TVu20} and \prompttuning~\cite{TVu22}, and has been successfully applied to the zero-shot cross-lingual transfer setting~\cite{JPhang20,KMaurya21} for \modeltuning. In Appendix~\ref{appendix:intermediate_tuning}, we show that intermediate tuning does not give reliable gains for \ssc{XGen}.

\begin{figure}
\begin{center}
\includegraphics[width=0.48\textwidth]{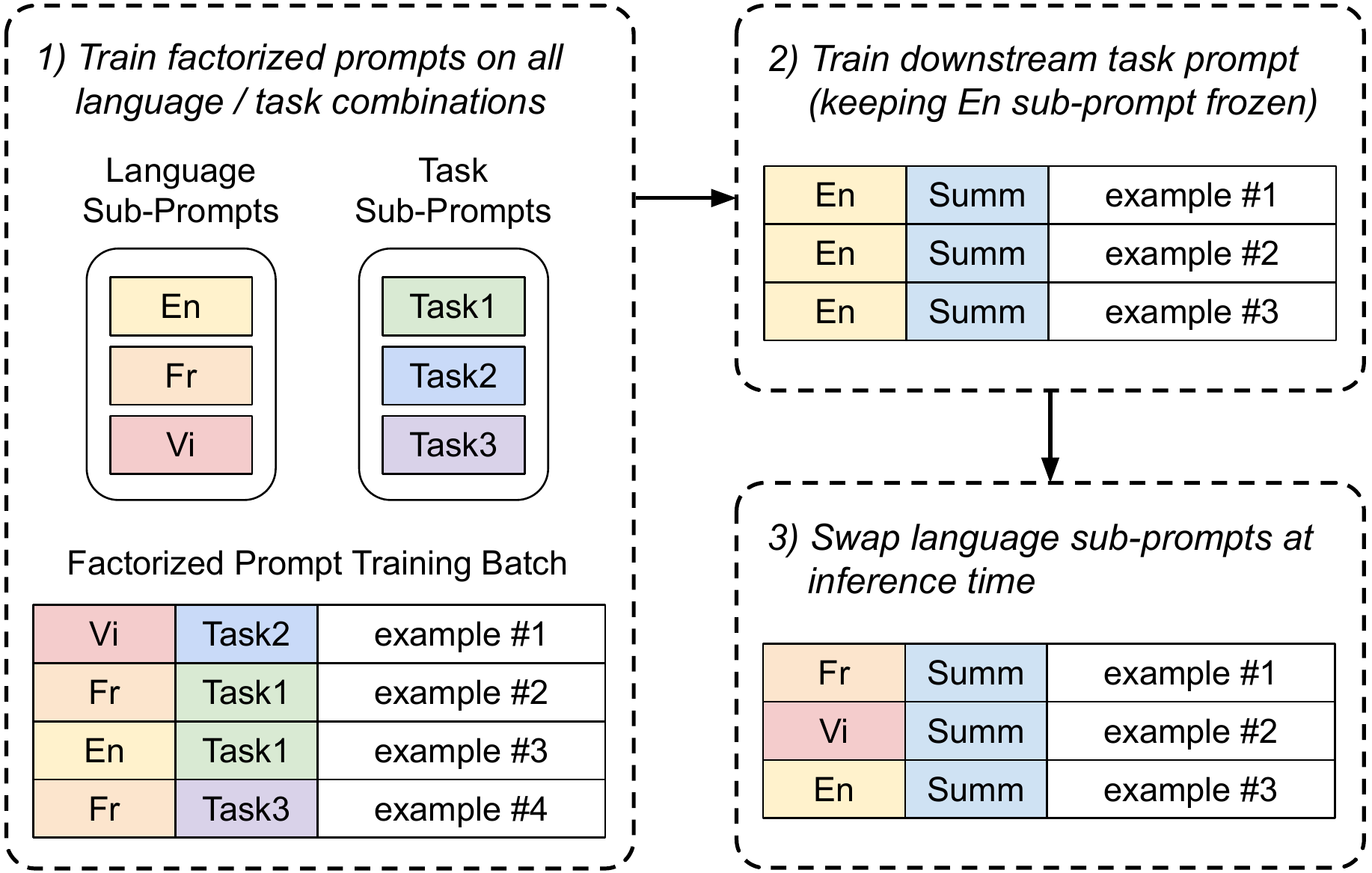}
\caption{Our ``factorized prompts'' approach learns recomposable language and task sub-prompts by training on all language / task combinations from a set of unsupervised tasks covering all target languages.}
\label{fig:factorized-prompts}
\end{center}
\end{figure}

\paragraph{Factorized prompts:} Inspired by the \ssc{MAD-X}\cite{JPfeiffer20} adapter-based framework that learns modular language and task representations to adapt a multilingual model to arbitrary tasks and languages, we propose a novel method, dubbed ``factorized prompts'' (\bssc{FP}) and specifically designed for \prompttuning. We attempt to %
decompose a soft prompt into ``task'' and ``language'' components that can be recombined in novel pairings (see Figure \ref{fig:factorized-prompts}) with the goal of learning soft prompts that consist of disentangled and interpretable components. Unlike \ssc{MAD-X}, which learns language and task adapters separately for each language and each task, we learn language and task sub-prompts jointly for all languages and tasks. %
While we do not actively incentivize disentanglement, our multi-task multilingual pretraining procedure encourages the general language and task-specific knowledge to be stored in separate regions of the prompt. Intuitively, we vary languages while keeping the task sub-prompt fixed to train one side of the prompt, and vary tasks while keeping the language sub-prompt fixed to learn the other side.

We use \mcfour data for all 18 \wikilinguazero languages to create 7 unsupervised tasks per language. We randomly initialize language and task sub-prompts, each $50$ tokens long.
For each training example in our multi-task multilingual mixture, the relevant task and language sub-prompts are concatenated to form a full $100$-token prompt. This training yields a set of learned language and task sub-prompts.\footnote{As our mixture of tasks is large, we tuned for $200{,}000$ steps for this training procedure.}  Next, we train a new task sub-prompt on \wikilinguazero English summarization while using a frozen copy of the English language sub-prompt.
Finally, when performing inference in another language, we replace the English sub-prompt with the target language sub-prompt, while continuing to use the learned summarization sub-prompt. To ablate the impact of the target language sub-prompt, we also report the performance using the English sub-prompt for all languages (\bssc{FP-En}).

We use 7 unsupervised tasks per language, including: the \ssc{prefix LM}, \ssc{span corruption}, and \ssc{i.i.d.~denoising} tasks described in~\citet{CRaffel20}; \ssc{LM}, the causal left-to-right \ssc{LM} task with no context provided, i.e., the encoder's input is empty; \ssc{missing prefix prediction}, predicting a missing prefix from the input; \ssc{n-token prefix prediction}, copying the first $n$-tokens of the input; and \ssc{missing n-token prefix prediction}, predicting the missing $n$-token prefix of the input. When training on \wikilinguazero, we initialize the task sub-prompt with the learned \ssc{span corruption} task sub-prompt.

To confirm that language-specific prompts trained in this way encode meaningful differences between languages, we visualize a clustered heatmap of the cosine similarities between prompts trained on a classic LM task for each language in mC4. We observe meaningful clusters reflecting both linguistic and geographical similarities across languages. See Appendix \ref{appendix:clustering} for details.

\begin{figure*}[t!]
\centering
\includegraphics[width=0.85\textwidth]{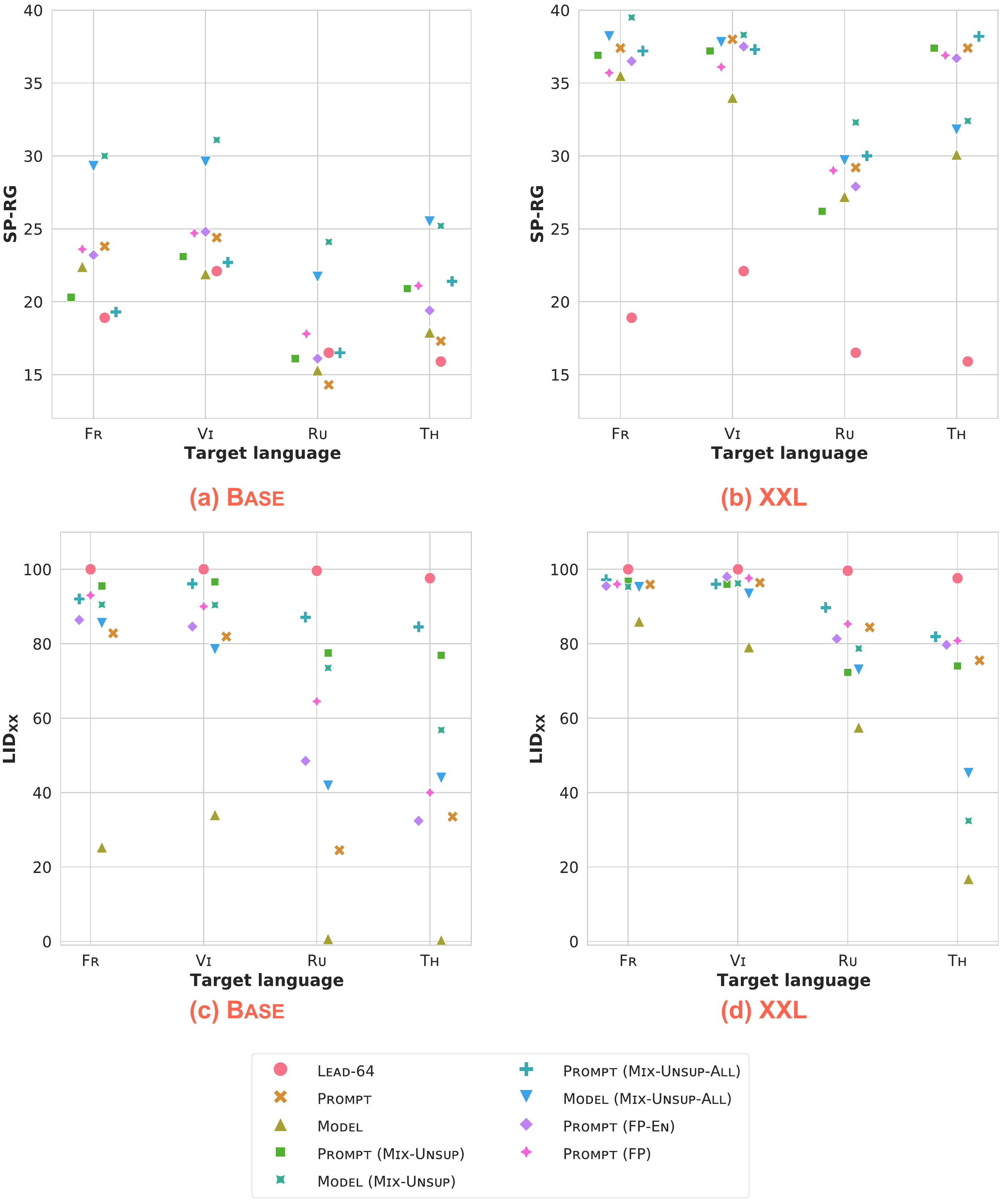}
\caption{\ssc{SP-Rouge} (top) and language accuracy (bottom) performance at \ssc{Base} and \ssc{XXL} sizes of our proposed approaches: mixing unsupervised data (\ssc{Mix}), and factorized prompts (\ssc{FP}). See  Appendix~\ref{appendix:our_methods} for full results.}
\label{figure:our_methods}
\vspace{-2mm}
\end{figure*}

\subsection{Results and Discussion}

\paragraph{Mixing in multilingual data prevents catastrophic forgetting:} In Figure~\ref{figure:our_methods}, we observe that mixing in unsupervised multilingual data helps prevent catastrophic forgetting in all conditions, increasing the likelihood of predicting text in the target language. With \modeltuning, this improved language accuracy reliably translates into higher end task performance (\ssc{SP-Rouge}). For \prompttuning, mixing provides gains for non-Latin script languages (\ssc{Ru} and \ssc{Th}) where catastrophic forgetting is more severe; for Latin-script languages (\ssc{Fr} and \ssc{Vi}), mixing harms the overall summarization quality, despite achieving higher language accuracy.%

Mixing in multilingual data in \emph{all} \ssc{Wikilingua} languages leads to similar results, with a marginal drop in performance. Thus, if the desired target language is known ahead of time, the simpler strategy of mixing in just that language should be preferred. However, in cases where the inference language is unknown, mixing many languages is also effective.%

\paragraph{Factorized prompts are helpful for overcoming severe catastrophic forgetting:} Factorized prompts are successful at improving target language accuracy in all conditions. However, this does not always translate to higher \ssc{SP-Rouge}. When language accuracy is already relatively high (for Latin-script languages, and for \ssc{XXL} models), factorized prompts are not helpful. However, in settings where vanilla \prompttuning shows the most severe forgetting (e.g., at \ssc{Base} size, on non-Latin script languages), factorized prompts provide large gains, similar to or exceeding our mixing approach.%

\section{Qualitative Analysis}
\label{section:qualitative_analysis}

\begin{table}
\begin{center}
\includegraphics[width=0.48\textwidth]{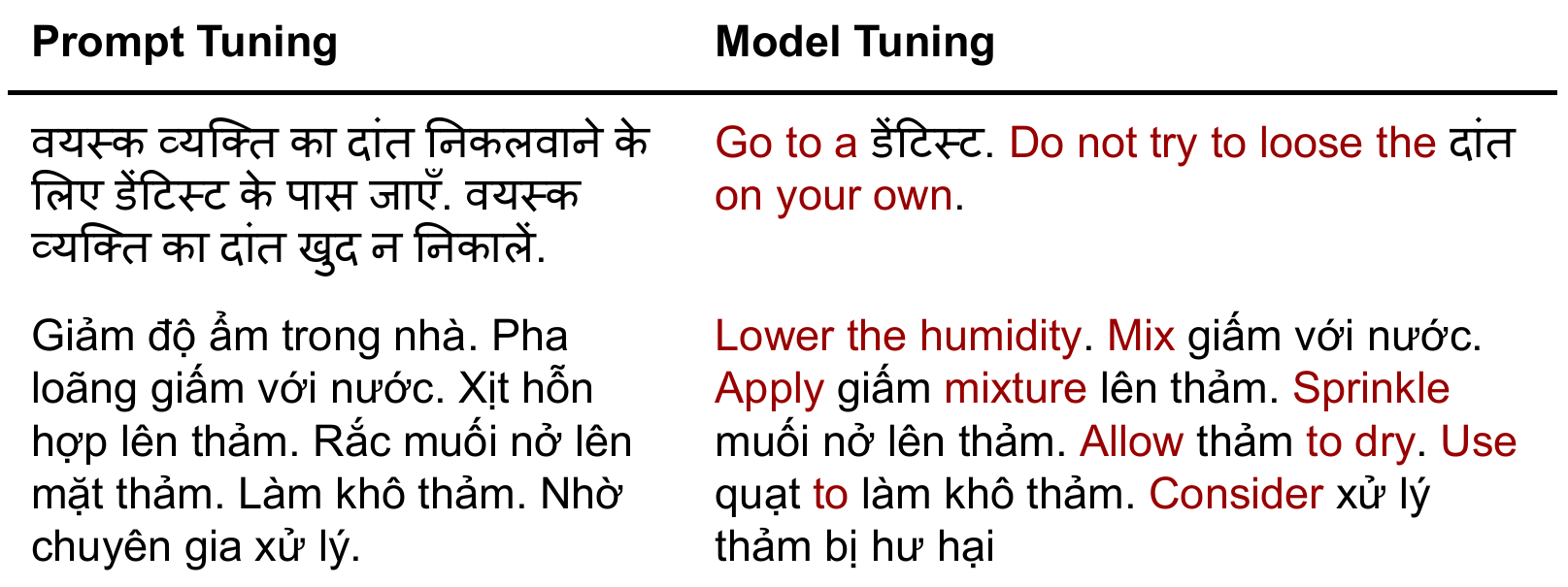}
\caption{Sample Hindi (top) and Vietnamese (bottom) predictions of our \ssc{XXL} model tuned with \prompttuning and \modeltuning. While the summaries are all understandable to a bilingual speaker, \prompttuning tends to stay within the target language, whereas \modeltuning is more prone to code switching between English (red) and the target language.}
\label{tab:example_predictions}
\end{center}
\end{table}

To better understand qualitative differences between the solutions reached by \modeltuning and \prompttuning, two authors who were native speakers of Vietnamese and Hindi inspected 50 predictions of each method at the \ssc{XXL} model size.

For both languages, we observed that the \modeltuning predictions were much more likely to include ``code-switching'', alternating between English and the target language, sometimes several times within a single sentence, as seen in Table~\ref{tab:example_predictions}. By comparison, the \prompttuning predictions were more likely to use a consistent language throughout---typically staying entirely within the target language, but for some predictions resorting entirely to English. For both methods and both languages, we found code-switching predictions to generally be well-formed, in the sense that a bilingual speaker could extract the intended meaning, and that it served as a reasonable summary.

For Hindi, the \prompttuning method showed lower mean \ssc{SP-Rouge} scores than \modeltuning ($17.9$ vs.~$23.1$), and had higher variances across runs (std: $5.1$ vs.~$0.7$). Manual inspection showed that the lower-scoring \prompttuning runs had far more predictions that were entirely English, explaining the lower \ssc{SP-Rouge} scores.

For Vietnamese, \prompttuning achieved higher \ssc{SP-Rouge} than \modeltuning ($38.0$ vs.~$34.0$), with low variance in both cases (std: $\le0.5$). On inspection, we found that most \prompttuning predictions were entirely in Vietnamese, whereas \modeltuning predictions typically contained at least some English. The \prompttuning summaries tended to be shorter, but were often judged to be as good or better than the ground truth summaries. The \modeltuning summaries tended to be a bit longer. If mentally translating any English words back to Vietnamese, the quality was judged to be similar to the prompt tuning summaries, suggesting that the lower \ssc{SP-Rouge} score is primarily due to the presence of intervening English.

\section{Related Work}
\label{section:related_work}

Mixing unlabeled multilingual data in during fine-tuning can be viewed a version of rehearsal \cite{ARobins95}, commonly used to mitigate catastrophic forgetting. Related work has used this mixing \cite{LXue21,SShakeri21} to combat ``accidental translation'', a symptom of English overfitting. However, these works are concerned with \modeltuning, whereas we apply it to \prompttuning. Other methods of combatting catastrophic forgetting include the slowing (or stopping) of updates for some parameters. \citet{JKirkpatrick17} reduce the learning rate of parameters important for earlier tasks as they train on new ones. \citet{KMaurya21} similarly stop learning for some parameters by only training input and output layers. In the context of prompt tuning, \citet{CQin22} address catastrophic forgetting during continual learning of new domains by combining the new training data with pseudo-labeled data of previous domains.

Previous work has also explored intermediate adaptation of pre-trained models, which has been shown to be effective for \modeltuning \cite{JHoward18,JPhang19,TVu20,TVu21} and \prompttuning \cite{TVu22}. \citet{JPhang20} apply intermediate adaptation in the multilingual domain, but use English in the adaption instead of the target language. \citet{KMaurya21} use a cross-lingual intermediate task. Unlike our task, theirs is designed to closely match the downstream task. Several works use intermediate adaptation to create a model that is better in the zero- or few-shot settings \cite{JWei22,VSanh22,SMin22}, but these target generalization to new tasks, whereas we focus on generalizing to new languages within one task.

Many parameter-efficient adaption methods exist \cite{SRebuffi17,NHoulsby19,RMahabadi21b,EZaken21,EHu22} and some have shown strong performance under domain shift \cite{BLester21,XLi21}. We chose \prompttuning due to its simplicity and the localization of parameters---making the implementation of factorized prompts easy. See \citet{PLiu21}, \citet{JHe22}, and \citet{HLiu22} for detailed discussion of the differences between these methods.

Other work explores cross-lingual transfer learning with parameter-efficient methods. \citet{MZhao21} find that soft prompts can effectively be used in cross-lingual settings, but their work is constrained to classification. \citet{JPfeiffer20} use adapters rather than prompts and leverage parameter-efficient learning to create separate language and task understanding modules that can be combined at inference time.%

There has been recent interest in cross-lingual generation. \citet{KMaurya21} and \citet{ZChi20} evaluate their methods using cross-lingual generation, including summarization as we do. However, \citet{ZChi20} use parallel data during pre-training to ``align'' representations across languages during pre-training while our approach does not. %

\section{Conclusion}
\label{section:conclusion}

In this work, we explored how different adaptation methods fare on the challenging ``\ssc{XGen}'' task of zero-shot cross-lingual summarization. While many methods struggled with catastrophic forgetting (outputting English rather than the target language), we observed two factors helped to mitigate this problem: (1) increasing model scale, and (2) decreasing the number of parameters tuned during adaptation. 
When all of a model's weights are tuned on English (\modeltuning), forgetting is quick and severe. By contrast, limiting the tunable parameters to a smaller soft prompt (\prompttuning) helps to combat forgetting, though prompt size is an important variable to control.

To further close the gap with supervised methods, we explored two adaptation techniques---one entirely novel, and one that has been used before, but not in combination with parameter-efficient methods like \prompttuning. We find that mixing in unsupervised multilingual data is always helpful. 
Our novel approach, ``factorized prompts'', is helpful at smaller model sizes, but has no benefit at larger sizes. We hope that future work will continue to explore \ssc{XGen} tasks including \wikilinguazero, and develop stronger zero-shot adaptation techniques to allow multilingual models to reliably generate coherent text in any target language.
\section{Limitations}

Our work focuses on a single \ssc{XGen} task, \mbox{\ssc{WikiLingua-0}} summarization. In future work, it would be valuable to see if our findings generalize to additional domains and tasks, including those beyond summarization.

\ssc{WikiLingua-0} is not a traditional summarization task. Rather than news articles, the input documents are how-to guides, and the summaries are ``headings'' for each step, which may be more terse than a traditional summary. We observed some minor data quality issues in \mbox{\ssc{WikiLingua-0}}, including \fssc{HTML} code present in some target strings, and artifacts of machine translation evident in some non-English documents. Nevertheless, we believe that \mbox{\ssc{WikiLingua-0}} is a meaningful and challenging \ssc{XGen} task, with the notable advantage of covering a range of high- and low-resource languages from diverse language families and with diverse scripts.

In evaluating parameter-efficient methods, we focused on \prompttuning due to its simplicity. There are a growing number of other parameter-efficient methods that could also be tested, including \ssc{Adapters}~\cite{SRebuffi17, NHoulsby19}, \ssc{BitFiT}~\cite{EZaken21}, \ssc{Prefix-Tuning}~\cite{XLi21}, \ssc{(IA)$^3$}~\cite{HLiu22}, and many more; see \citet{PLiu21}, \citet{JHe22}, and \citet{HLiu22} for detailed discussion of the differences between these methods. We expect many of the benefits of tuning fewer parameters to persist across methods, but this remains to be explored.

\section*{Acknowledgements}
We thank Thibault Sellam, Sebastian Gehrmann, Kalpesh Krishna, Marzena Karpinska, and the members of the UMass NLP group for helpful discussion and feedback. We would also like to thank Grady Simon, Xavier Garcia, and Douglas Eck for their comments on this manuscript. Vu and Iyyer are partially supported by awards IIS-1955567 and IIS-2046248	from the National Science Foundation (NSF).

\bibliography{custom}
\bibliographystyle{acl_natbib}

\clearpage
\newpage
\appendix
\section*{Appendices}

\label{section:appendices}
\section{Evaluation on zero-shot cross-lingual benchmarks}
\label{appendix:xlingeval}
From Table~\ref{tbl:xnli} to Table~\ref{tbl:wikilingua}, we show our results for \modeltuning and \prompttuning across different zero-shot cross-lingual benchmarks. Overall, we find that \ssc{ModelTuning} typically performs better than \ssc{PromptTuning}, although \ssc{PromptTuning} at scale (i.e., \ssc{XXL}) matches the performance of \ssc{ModelTuning} on English and can yield better results on some languages.
\clearpage
\newpage

\begin{table*}[t!]
\centering
\begin{adjustbox}{max width=\textwidth}
\begin{tabular}{c c c c c c c c c c c c c c c c c}
\toprule
\multirow{2}{*}{\textbf{Size}} & \multirow{2}{*}{\textbf{Method}} & \multicolumn{15}{c}{\textbf{Language}} \\
\cmidrule(l){3-17} & & en & ar & bg & de & el & es & fr & hi & ru & sw & th & tr & ur & vi & zh \\
\midrule
\multirow{2}{*}{\ssc{Base}} & \prompt & 78.8$_{\smallsup{1.2}}$ & 64.7$_{\smallsup{0.4}}$ & 68.9$_{\smallsup{0.5}}$ & 68.4$_{\smallsup{1.0}}$ & 70.1$_{\smallsup{0.8}}$ & 73.7$_{\smallsup{0.5}}$ & 75.6$_{\smallsup{1.1}}$ & 65.1$_{\smallsup{0.4}}$ & 68.0$_{\smallsup{0.6}}$ & 62.5$_{\smallsup{0.2}}$ & 69.7$_{\smallsup{0.9}}$ & 67.6$_{\smallsup{0.3}}$ & 60.9$_{\smallsup{0.7}}$ & 70.7$_{\smallsup{1.5}}$ & 70.3$_{\smallsup{1.3}}$ \\
& \model & 87.1$_{\smallsup{0.2}}$ & 72.3$_{\smallsup{0.2}}$ & 78.4$_{\smallsup{0.7}}$ & 77.7$_{\smallsup{0.2}}$ & 82.0$_{\smallsup{1.0}}$ & 84.5$_{\smallsup{0.8}}$ & 80.8$_{\smallsup{0.6}}$ & 70.3$_{\smallsup{1.1}}$ & 74.8$_{\smallsup{0.7}}$ & 69.3$_{\smallsup{1.0}}$ & 74.3$_{\smallsup{1.0}}$ & 73.2$_{\smallsup{1.0}}$ & 68.0$_{\smallsup{0.3}}$ & 77.7$_{\smallsup{0.5}}$ & 72.9$_{\smallsup{1.0}}$ \\
& $\Delta_{\text{P-M}}$ & -8.3 & -7.6 & -9.5 & -9.3 & -11.9 & -10.8 & -5.2 & -5.2 & -6.8 & -6.8 & -4.6 & -5.6 & -7.1 & -7.0 & -2.6 \\
\midrule
\multirow{2}{*}{\ssc{XXL}} & \prompt & 91.5$_{\smallsup{0.2}}$ & 81.5$_{\smallsup{0.2}}$ & 87.1$_{\smallsup{0.4}}$ & 88.5$_{\smallsup{0.4}}$ & 88.9$_{\smallsup{0.8}}$ & 90.1$_{\smallsup{0.4}}$ & 88.4$_{\smallsup{1.1}}$ & 84.5$_{\smallsup{0.4}}$ & 83.3$_{\smallsup{0.4}}$ & 80.7$_{\smallsup{0.7}}$ & 81.6$_{\smallsup{0.3}}$ & 83.7$_{\smallsup{0.4}}$ & 78.9$_{\smallsup{0.4}}$ & 85.1$_{\smallsup{1.0}}$ & 83.7$_{\smallsup{0.4}}$ \\
& \model & 92.8$_{\smallsup{0.6}}$ & 85.6$_{\smallsup{0.6}}$ & 89.3$_{\smallsup{0.5}}$ & 89.2$_{\smallsup{0.3}}$ & 89.5$_{\smallsup{0.8}}$ & 90.8$_{\smallsup{0.0}}$ & 88.5$_{\smallsup{0.8}}$ & 84.5$_{\smallsup{0.5}}$ & 82.9$_{\smallsup{0.8}}$ & 83.7$_{\smallsup{0.7}}$ & 78.8$_{\smallsup{0.9}}$ & 83.3$_{\smallsup{1.0}}$ & 81.5$_{\smallsup{0.7}}$ & 87.6$_{\smallsup{0.9}}$ & 84.1$_{\smallsup{0.2}}$ \\
& $\Delta_{\text{P-M}}$ & -1.3 & -4.1 & -2.2 & -0.7 & -0.6 & -0.7 & -0.1 & \textcolor{mygreen}{0.0} & \textcolor{mygreen}{0.4} & -3.0 & \textcolor{mygreen}{2.8} & \textcolor{mygreen}{0.4} & -2.6 & -2.5 & -0.4 \\
\bottomrule
\end{tabular}
\end{adjustbox}
\caption{Best validation accuracy per language on \ssc{XNLI}.}
\label{tbl:xnli}
\end{table*}

\begin{table*}[t!]
\centering
\begin{adjustbox}{max width=\textwidth}
\begin{tabular}{c c c c c c c c c c c c c c}
\toprule
\multirow{2}{*}{\textbf{Size}} & \multirow{2}{*}{\textbf{Method}} & \multicolumn{11}{c}{\textbf{Language}} \\
\cmidrule(l){3-13} & & en & ar & de & el & es & hi & ru & th & tr & vi & zh \\
\midrule
\multirow{2}{*}{\ssc{Base}} & \prompt & 83.9$_{\smallsup{0.3}}$ & 63.0$_{\smallsup{1.5}}$& 70.7$_{\smallsup{0.8}}$& 63.5$_{\smallsup{0.7}}$& 75.6$_{\smallsup{1.1}}$& 61.4$_{\smallsup{1.7}}$& 61.6$_{\smallsup{1.4}}$& 58.3$_{\smallsup{1.0}}$& 60.9$_{\smallsup{0.8}}$& 68.7$_{\smallsup{0.8}}$& 45.3$_{\smallsup{1.2}}$ \\ 
& \model & 91.9$_{\smallsup{0.3}}$ & 72.9$_{\smallsup{0.9}}$& 76.9$_{\smallsup{0.7}}$& 68.4$_{\smallsup{0.4}}$& 84.9$_{\smallsup{0.7}}$& 67.5$_{\smallsup{0.9}}$& 69.8$_{\smallsup{1.0}}$& 63.4$_{\smallsup{1.1}}$& 69.3$_{\smallsup{0.9}}$& 77.2$_{\smallsup{0.2}}$& 53.3$_{\smallsup{0.4}}$ \\
& $\Delta_{\text{P-M}}$ & -8.0 & -9.9 & -6.2 & -4.9 & -9.3 & -6.1 & -8.2 & -5.1 & -8.4 & -8.5 & -8.0 \\
\midrule
\multirow{2}{*}{\ssc{XXL}} & \prompt & 95.0$_{\smallsup{0.1}}$ & 83.6$_{\smallsup{0.3}}$& 84.9$_{\smallsup{0.9}}$& 76.6$_{\smallsup{0.6}}$& 92.5$_{\smallsup{0.5}}$& 77.7$_{\smallsup{1.1}}$& 80.3$_{\smallsup{0.6}}$& 71.6$_{\smallsup{1.5}}$& 81.9$_{\smallsup{0.5}}$& 85.5$_{\smallsup{0.2}}$& 60.8$_{\smallsup{0.7}}$ \\
& \model & 95.5$_{\smallsup{0.2}}$ & 88.6$_{\smallsup{0.1}}$& 86.3$_{\smallsup{0.9}}$& 81.8$_{\smallsup{0.7}}$& 92.4$_{\smallsup{0.4}}$& 82.1$_{\smallsup{0.8}}$& 85.0$_{\smallsup{0.5}}$& 75.8$_{\smallsup{0.8}}$& 84.6$_{\smallsup{0.2}}$& 88.5$_{\smallsup{0.5}}$& 64.9$_{\smallsup{0.8}}$ \\
& $\Delta_{\text{P-M}}$ & -0.5 & -5.0 & -1.4 & -5.2 & \textcolor{mygreen}{0.1} & -4.4 & -4.7 & -4.2 & -2.7 & -3.0 & -4.1 \\
\bottomrule
\end{tabular}
\end{adjustbox}
\caption{Best validation \ssc{F1} per language on \ssc{XQuAD}.}
\label{tbl:xquad}
\end{table*}

\clearpage
\newpage

\begin{table*}[t!]
\centering
\begin{adjustbox}{max width=\textwidth}
\begin{tabular}{c c c c c c c c c}
\toprule
\multirow{2}{*}{\textbf{Size}} & \multirow{2}{*}{\textbf{Method}} & \multicolumn{7}{c}{\textbf{Language}} \\
\cmidrule(l){3-9} & & en & ar & de & es & hi & vi & zh \\
\midrule
\multirow{2}{*}{\ssc{Base}} & \prompt & 75.5$_{\smallsup{0.4}}$ & 45.1$_{\smallsup{1.3}}$ & 55.2$_{\smallsup{0.8}}$ & 63.0$_{\smallsup{1.0}}$ & 47.9$_{\smallsup{2.1}}$ & 53.8$_{\smallsup{0.6}}$ & 55.1$_{\smallsup{0.6}}$ \\
& \model & 79.4$_{\smallsup{0.4}}$ & 53.8$_{\smallsup{0.2}}$ & 62.8$_{\smallsup{0.4}}$ & 69.7$_{\smallsup{0.6}}$ & 55.7$_{\smallsup{0.3}}$ & 62.8$_{\smallsup{0.6}}$ & 62.7$_{\smallsup{0.5}}$ \\
& $\Delta_{\text{P-M}}$ & -3.9 & -8.7 & -7.6 & -6.7 & -7.8 & -9.0 & -7.6 \\
\midrule
\multirow{2}{*}{\ssc{XXL}} & \prompt & 85.4$_{\smallsup{0.4}}$ & 63.7$_{\smallsup{0.8}}$ & 72.0$_{\smallsup{0.8}}$ & 76.3$_{\smallsup{0.4}}$ & 68.4$_{\smallsup{0.5}}$ & 70.1$_{\smallsup{0.6}}$ & 71.0$_{\smallsup{0.4}}$ \\
& \model & 84.7$_{\smallsup{0.5}}$ & 71.1$_{\smallsup{0.5}}$ & 72.8$_{\smallsup{0.1}}$ & 79.0$_{\smallsup{0.2}}$ & 73.9$_{\smallsup{0.3}}$ & 71.4$_{\smallsup{0.3}}$ & 75.4$_{\smallsup{0.5}}$ \\
& $\Delta_{\text{P-M}}$ & \textcolor{mygreen}{0.7} & -7.4 & -0.8 & -2.7 & -5.5 & -1.3 & -4.4 \ \\
\bottomrule
\end{tabular}
\end{adjustbox}
\caption{Best validation \ssc{F1} per language on \ssc{MLQA}.}
\label{tbl:mlqa}
\end{table*}

\begin{table*}[t!]
\centering
\begin{adjustbox}{max width=\textwidth}
\begin{tabular}{c c c c c c c c c c c}
\toprule
\multirow{2}{*}{\textbf{Size}} & \multirow{2}{*}{\textbf{Method}} & \multicolumn{9}{c}{\textbf{Language}} \\
\cmidrule(l){3-11} & & en & ar & bn & fi & id & ko & ru & sw & te \\
\midrule
\multirow{2}{*}{\ssc{Base}} & \prompt & 68.1$_{\smallsup{1.5}}$ & 61.6$_{\smallsup{4.3}}$ & 37.2$_{\smallsup{0.5}}$ & 56.6$_{\smallsup{2.0}}$ & 59.6$_{\smallsup{3.4}}$ & 35.3$_{\smallsup{0.7}}$ & 58.2$_{\smallsup{1.0}}$ & 46.7$_{\smallsup{3.1}}$ & 41.1$_{\smallsup{3.3}}$ \\
& \model & 71.5$_{\smallsup{0.9}}$ & 69.9$_{\smallsup{1.2}}$ & 41.5$_{\smallsup{1.3}}$ & 67.6$_{\smallsup{0.7}}$ & 77.5$_{\smallsup{0.6}}$ & 48.8$_{\smallsup{0.4}}$ & 57.8$_{\smallsup{1.1}}$ & 61.2$_{\smallsup{1.0}}$ & 48.1$_{\smallsup{1.5}}$ \\
& $\Delta_{\text{P-M}}$ & -3.4 & -8.3 & -4.3 & -11.0 & -17.9 & -13.5 & \textcolor{mygreen}{0.4} & -14.5 & -7.0 \\
\midrule
\multirow{2}{*}{\ssc{XXL}} & \prompt & 82.8$_{\smallsup{0.5}}$ & 78.1$_{\smallsup{0.6}}$ & 73.9$_{\smallsup{1.4}}$ & 76.6$_{\smallsup{0.8}}$ & 83.9$_{\smallsup{0.4}}$ & 73.7$_{\smallsup{2.0}}$ & 69.7$_{\smallsup{0.7}}$ & 71.8$_{\smallsup{2.3}}$ & 77.9$_{\smallsup{1.1}}$ \\
& \model & 85.1$_{\smallsup{0.4}}$ & 85.7$_{\smallsup{0.1}}$ & 83.4$_{\smallsup{1.5}}$ & 82.3$_{\smallsup{0.5}}$ & 88.7$_{\smallsup{0.3}}$ & 76.3$_{\smallsup{0.5}}$ & 76.5$_{\smallsup{1.1}}$ & 82.9$_{\smallsup{0.6}}$ & 79.7$_{\smallsup{0.1}}$ \\
& $\Delta_{\text{P-M}}$ & -2.3 & -7.6 & -9.5 & -5.7 & -4.8 & -2.6 & -6.8 & -11.1 & -1.8 \ \\
\bottomrule
\end{tabular}
\end{adjustbox}
\caption{Best validation \ssc{F1} per language on \ssc{TyDiQA}.}
\label{tbl:tydiqa}
\end{table*}

\clearpage
\newpage

\begin{table*}[t!]
\centering
\begin{adjustbox}{max width=\textwidth}
\begin{tabular}{c c c c c c c c c}
\toprule
\multirow{2}{*}{\textbf{Size}} & \multirow{2}{*}{\textbf{Method}} & \multicolumn{7}{c}{\textbf{Language}} \\
\cmidrule(l){3-9} & & en & de & es & fr & ja & ko & zh \\
\midrule
\multirow{2}{*}{\ssc{Base}} & \prompt & 94.3$_{\smallsup{0.8}}$ & 85.3$_{\smallsup{0.5}}$ & 88.1$_{\smallsup{0.8}}$ & 88.9$_{\smallsup{0.8}}$ & 80.8$_{\smallsup{0.3}}$ & 79.7$_{\smallsup{0.8}}$ & 82.3$_{\smallsup{1.1}}$ \\
& \model & 94.9$_{\smallsup{0.7}}$ & 89.1$_{\smallsup{0.4}}$ & 90.8$_{\smallsup{0.3}}$ & 90.7$_{\smallsup{0.5}}$ & 84.1$_{\smallsup{0.8}}$ & 83.5$_{\smallsup{0.9}}$ & 84.3$_{\smallsup{0.5}}$ \\
& $\Delta_{\text{P-M}}$ & -0.6 & -3.8 & -2.7 & -1.8 & -3.3 & -3.8 & -2.0 \\
\midrule
\multirow{2}{*}{\ssc{XXL}} & \prompt & 96.8$_{\smallsup{0.6}}$ & 90.7$_{\smallsup{0.2}}$ & 92.9$_{\smallsup{0.4}}$ & 93.5$_{\smallsup{0.4}}$ & 88.1$_{\smallsup{0.4}}$ & 86.0$_{\smallsup{0.9}}$ & 88.8$_{\smallsup{1.0}}$ \\
& \model & 96.4$_{\smallsup{0.3}}$ & 91.9$_{\smallsup{0.2}}$ & 92.8$_{\smallsup{0.3}}$ & 93.9$_{\smallsup{0.5}}$ & 87.2$_{\smallsup{1.2}}$ & 89.6$_{\smallsup{0.3}}$ & 91.9$_{\smallsup{0.4}}$ \\
& $\Delta_{\text{P-M}}$ & \textcolor{mygreen}{0.4} & -1.2 & \textcolor{mygreen}{0.1} & -0.4 & \textcolor{mygreen}{0.9} & -3.6 & -3.1 \ \\
\bottomrule
\end{tabular}
\end{adjustbox}
\caption{Best validation accuracy per language on \ssc{PAWS-X}.}
\label{tbl:pawsx}
\end{table*}

\begin{table*}[t!]
\centering
\begin{adjustbox}{max width=\textwidth}
\begin{tabular}{c c c c c c c c c c c c c c c c c c c c c c}
\toprule
\multirow{2}{*}{\textbf{Size}} & \multirow{2}{*}{\textbf{Method}} & \multicolumn{20}{c}{\textbf{Language}} \\
\cmidrule(l){3-22} & & en & af & ar & bg & bn & de & el & es & et & eu & fa & fi & fr & he & hi & hu & id & it & ja & jv \\
\midrule
\multirow{2}{*}{\ssc{Base}} & \prompt & 83.3$_{\smallsup{0.4}}$ & 73.3$_{\smallsup{1.3}}$ & 43.6$_{\smallsup{1.0}}$ & 74.8$_{\smallsup{0.7}}$ & 64.5$_{\smallsup{0.8}}$ & 68.9$_{\smallsup{1.0}}$ & 68.2$_{\smallsup{1.0}}$ & 74.3$_{\smallsup{1.1}}$ & 62.5$_{\smallsup{2.3}}$ & 46.2$_{\smallsup{2.4}}$ & 31.9$_{\smallsup{1.8}}$ & 63.2$_{\smallsup{0.4}}$ & 78.3$_{\smallsup{0.1}}$ & 48.9$_{\smallsup{1.3}}$ & 65.8$_{\smallsup{2.9}}$ & 68.4$_{\smallsup{0.3}}$ & 54.5$_{\smallsup{1.0}}$ & 81.9$_{\smallsup{1.2}}$ & 34.9$_{\smallsup{0.9}}$ & 53.7$_{\smallsup{0.7}}$ \\
& \model & 87.3$_{\smallsup{0.8}}$ & 72.3$_{\smallsup{1.5}}$ & 52.1$_{\smallsup{2.4}}$ & 63.2$_{\smallsup{2.1}}$ & 72.0$_{\smallsup{1.3}}$ & 64.5$_{\smallsup{0.6}}$ & 59.2$_{\smallsup{1.7}}$ & 66.7$_{\smallsup{1.3}}$ & 68.6$_{\smallsup{1.2}}$ & 49.1$_{\smallsup{1.8}}$ & 30.4$_{\smallsup{1.9}}$ & 71.5$_{\smallsup{0.3}}$ & 79.5$_{\smallsup{0.3}}$ & 44.4$_{\smallsup{1.0}}$ & 65.7$_{\smallsup{0.9}}$ & 66.8$_{\smallsup{0.4}}$ & 51.6$_{\smallsup{0.2}}$ & 82.5$_{\smallsup{0.7}}$ & 35.7$_{\smallsup{1.4}}$ & 46.5$_{\smallsup{1.2}}$ \\
& $\Delta_{\text{P-M}}$ & -4.0 & \textcolor{mygreen}{1.0} & -8.5 & \textcolor{mygreen}{11.6} & -7.5 & \textcolor{mygreen}{4.4} & \textcolor{mygreen}{9.0} & \textcolor{mygreen}{7.6} & -6.1 & -2.9 & \textcolor{mygreen}{1.5} & -8.3 & -1.2 & \textcolor{mygreen}{4.5} & \textcolor{mygreen}{0.1} & \textcolor{mygreen}{1.6} & \textcolor{mygreen}{2.9} & -0.6 & -0.8 & \textcolor{mygreen}{7.2} \\
\midrule
\multirow{2}{*}{\ssc{XXL}} & \prompt & 91.5$_{\smallsup{0.4}}$ & 83.3$_{\smallsup{0.8}}$ & 51.0$_{\smallsup{0.8}}$ & 84.1$_{\smallsup{1.0}}$ & 79.1$_{\smallsup{1.2}}$ & 77.4$_{\smallsup{0.3}}$ & 79.3$_{\smallsup{0.8}}$ & 83.4$_{\smallsup{1.1}}$ & 78.4$_{\smallsup{1.7}}$ & 67.2$_{\smallsup{1.8}}$ & 41.0$_{\smallsup{1.9}}$ & 77.1$_{\smallsup{2.6}}$ & 86.6$_{\smallsup{1.2}}$ & 61.2$_{\smallsup{0.3}}$ & 75.4$_{\smallsup{2.1}}$ & 79.9$_{\smallsup{0.9}}$ & 71.1$_{\smallsup{4.1}}$ & 89.0$_{\smallsup{1.2}}$ & 41.8$_{\smallsup{1.7}}$ & 71.8$_{\smallsup{0.5}}$ \\
& \model & 91.7$_{\smallsup{0.4}}$ & 82.1$_{\smallsup{1.1}}$ & 62.0$_{\smallsup{2.3}}$ & 86.6$_{\smallsup{0.7}}$ & 82.7$_{\smallsup{0.5}}$ & 79.3$_{\smallsup{0.6}}$ & 79.6$_{\smallsup{0.7}}$ & 83.0$_{\smallsup{0.2}}$ & 74.7$_{\smallsup{1.1}}$ & 57.2$_{\smallsup{1.9}}$ & 52.5$_{\smallsup{0.9}}$ & 69.5$_{\smallsup{1.1}}$ & 88.5$_{\smallsup{0.7}}$ & 60.5$_{\smallsup{0.7}}$ & 81.1$_{\smallsup{0.3}}$ & 74.1$_{\smallsup{0.4}}$ & 71.0$_{\smallsup{3.4}}$ & 88.5$_{\smallsup{0.5}}$ & 50.7$_{\smallsup{0.8}}$ & 65.7$_{\smallsup{0.7}}$ \\
& $\Delta_{\text{P-M}}$ & -0.2 & \textcolor{mygreen}{1.2} & -11.0 & -2.5 & -3.6 & -1.9 & -0.3 & \textcolor{mygreen}{0.4} & \textcolor{mygreen}{3.7} & \textcolor{mygreen}{10.0} & -11.5 & \textcolor{mygreen}{7.6} & -1.9 & \textcolor{mygreen}{0.7} & -5.7 & \textcolor{mygreen}{5.8} & \textcolor{mygreen}{0.1} & \textcolor{mygreen}{0.5} & -8.9 & \textcolor{mygreen}{6.1} \\
\end{tabular}
\end{adjustbox}
\bigskip\bigskip
\begin{adjustbox}{max width=\textwidth}
\begin{tabular}{c c c c c c c c c c c c c c c c c c c c c c}
\toprule
\multirow{2}{*}{\textbf{Size}} & \multirow{2}{*}{\textbf{Method}} & \multicolumn{20}{c}{\textbf{Language}} \\
\cmidrule(l){3-22} & & ka & kk & ko & ml & mr & ms & my & nl & pt & ru & sw & ta & te & th & tl & tr & ur & vi & yo & zh \\
\midrule
\multirow{2}{*}{\ssc{Base}} & \prompt & 56.0$_{\smallsup{1.4}}$ & 44.7$_{\smallsup{2.0}}$ & 33.0$_{\smallsup{1.0}}$ & 47.0$_{\smallsup{1.7}}$ & 39.4$_{\smallsup{1.5}}$ & 76.8$_{\smallsup{0.9}}$ & 27.6$_{\smallsup{2.0}}$ & 79.0$_{\smallsup{1.2}}$ & 76.3$_{\smallsup{0.7}}$ & 58.0$_{\smallsup{1.6}}$ & 62.2$_{\smallsup{1.3}}$ & 45.6$_{\smallsup{2.5}}$ & 47.8$_{\smallsup{3.2}}$ & 10.9$_{\smallsup{0.1}}$ & 74.9$_{\smallsup{0.7}}$ & 68.3$_{\smallsup{0.8}}$ & 50.4$_{\smallsup{6.9}}$ & 69.6$_{\smallsup{0.7}}$ & 61.9$_{\smallsup{3.1}}$ & 33.9$_{\smallsup{3.0}}$ \\
& \model & 53.7$_{\smallsup{1.5}}$ & 20.7$_{\smallsup{1.9}}$ & 33.2$_{\smallsup{0.4}}$ & 45.1$_{\smallsup{0.5}}$ & 39.8$_{\smallsup{0.9}}$ & 75.4$_{\smallsup{0.8}}$ & 28.0$_{\smallsup{1.3}}$ & 80.2$_{\smallsup{2.3}}$ & 75.1$_{\smallsup{1.8}}$ & 50.3$_{\smallsup{1.2}}$ & 66.6$_{\smallsup{0.4}}$ & 43.2$_{\smallsup{0.7}}$ & 44.2$_{\smallsup{1.4}}$ & 9.9$_{\smallsup{0.7}}$ & 78.2$_{\smallsup{1.9}}$ & 60.3$_{\smallsup{1.6}}$ & 37.6$_{\smallsup{3.4}}$ & 74.8$_{\smallsup{1.8}}$ & 59.9$_{\smallsup{1.5}}$ & 41.0$_{\smallsup{1.8}}$ \\
& $\Delta_{\text{P-M}}$ & \textcolor{mygreen}{2.3} & \textcolor{mygreen}{24.0} & -0.2 & \textcolor{mygreen}{1.9} & -0.4 & \textcolor{mygreen}{1.4} & -0.4 & -1.2 & \textcolor{mygreen}{1.2} & \textcolor{mygreen}{7.7} & -4.4 & \textcolor{mygreen}{2.4} & \textcolor{mygreen}{3.6} & \textcolor{mygreen}{1.0} & -3.3 & \textcolor{mygreen}{8.0} & \textcolor{mygreen}{12.8} & -5.2 & \textcolor{mygreen}{2.0} & -7.1 \\
\midrule
\multirow{2}{*}{\ssc{XXL}} & \prompt & 70.5$_{\smallsup{2.5}}$ & 50.8$_{\smallsup{2.1}}$ & 51.2$_{\smallsup{1.4}}$ & 62.6$_{\smallsup{1.4}}$ & 57.2$_{\smallsup{3.8}}$ & 84.7$_{\smallsup{0.9}}$ & 42.5$_{\smallsup{1.8}}$ & 89.1$_{\smallsup{0.6}}$ & 86.9$_{\smallsup{0.9}}$ & 71.7$_{\smallsup{1.5}}$ & 77.8$_{\smallsup{0.7}}$ & 59.8$_{\smallsup{1.8}}$ & 57.8$_{\smallsup{0.1}}$ & 9.4$_{\smallsup{1.2}}$ & 83.3$_{\smallsup{1.6}}$ & 87.6$_{\smallsup{0.4}}$ & 81.7$_{\smallsup{3.0}}$ & 79.7$_{\smallsup{2.2}}$ & 60.3$_{\smallsup{0.5}}$ & 49.1$_{\smallsup{1.7}}$ \\
& \model & 71.9$_{\smallsup{1.1}}$ & 37.0$_{\smallsup{2.1}}$ & 46.1$_{\smallsup{0.6}}$ & 55.6$_{\smallsup{0.1}}$ & 54.8$_{\smallsup{0.6}}$ & 81.1$_{\smallsup{0.7}}$ & 38.5$_{\smallsup{0.8}}$ & 89.1$_{\smallsup{0.3}}$ & 87.4$_{\smallsup{0.7}}$ & 72.8$_{\smallsup{2.3}}$ & 78.3$_{\smallsup{0.5}}$ & 53.6$_{\smallsup{0.6}}$ & 53.6$_{\smallsup{0.9}}$ & 16.7$_{\smallsup{0.7}}$ & 84.4$_{\smallsup{0.6}}$ & 74.2$_{\smallsup{0.4}}$ & 82.8$_{\smallsup{0.7}}$ & 84.6$_{\smallsup{0.3}}$ & 68.2$_{\smallsup{3.1}}$ & 56.2$_{\smallsup{0.6}}$ \\
& $\Delta_{\text{P-M}}$ & -1.4 & \textcolor{mygreen}{13.8} & \textcolor{mygreen}{5.1} & \textcolor{mygreen}{7.0} & \textcolor{mygreen}{2.4} & \textcolor{mygreen}{3.6} & \textcolor{mygreen}{4.0} & \textcolor{mygreen}{0.0} & -0.5 & -1.1 & -0.5 & \textcolor{mygreen}{6.2} & \textcolor{mygreen}{4.2} & -7.3 & -1.1 & \textcolor{mygreen}{13.4} & -1.1 & -4.9 & -7.9 & -7.1 \ \\
\bottomrule
\end{tabular}
\end{adjustbox}
\vspace{-8.5mm}
\caption{Best validation span \ssc{F1} per language on \ssc{WikiAnn NER}.}
\label{tbl:wikiannner}
\end{table*}

\clearpage
\newpage

\begin{table*}[t!]
\centering
\begin{adjustbox}{max width=\textwidth}
\begin{tabular}{c c c c c c c c c c c c c c c c c c c c}
\toprule
\multirow{2}{*}{\textbf{Size}} & \multirow{2}{*}{\textbf{Method}} & \multicolumn{18}{c}{\textbf{Language}} \\

\cmidrule(l){3-20} & & en & es & pt & fr & de & ru & it & id & nl & ar & zh & vi & th & ja & ko & hi & cs & tr \\
\midrule
\multirow{2}{*}{\ssc{Small}} & \prompt & 24.5$_{\smallsup{0.2}}$& 20.2$_{\smallsup{0.6}}$& 20.7$_{\smallsup{0.3}}$& 19.2$_{\smallsup{0.1}}$& 15.4$_{\smallsup{0.3}}$& 11.4$_{\smallsup{0.1}}$& 18.3$_{\smallsup{0.5}}$& 19.0$_{\smallsup{0.6}}$& 16.9$_{\smallsup{0.2}}$& 16.0$_{\smallsup{0.3}}$& 12.8$_{\smallsup{0.5}}$& 21.4$_{\smallsup{0.2}}$& 14.9$_{\smallsup{0.4}}$& 12.1$_{\smallsup{0.1}}$& 14.8$_{\smallsup{0.3}}$& 11.2$_{\smallsup{0.6}}$& 14.0$_{\smallsup{0.1}}$& 13.7$_{\smallsup{0.0}}$ \\ 
& \model & 38.0$_{\smallsup{0.4}}$& 22.3$_{\smallsup{0.1}}$& 23.2$_{\smallsup{0.4}}$& 21.3$_{\smallsup{0.3}}$& 17.8$_{\smallsup{0.2}}$& 14.6$_{\smallsup{0.2}}$& 20.1$_{\smallsup{0.2}}$& 21.0$_{\smallsup{0.2}}$& 19.7$_{\smallsup{0.2}}$& 17.0$_{\smallsup{0.1}}$& 14.1$_{\smallsup{0.3}}$& 22.5$_{\smallsup{0.1}}$& 17.3$_{\smallsup{0.1}}$& 14.1$_{\smallsup{0.1}}$& 17.8$_{\smallsup{0.4}}$& 9.5$_{\smallsup{0.0}}$& 17.2$_{\smallsup{0.1}}$& 16.0$_{\smallsup{0.1}}$ \\
& $\Delta_{\text{P-M}}$ & -13.5 & -2.1 & -2.5 & -2.1 & -2.4 & -3.2 & -1.8 & -2.0 & -2.8 & -1.0 & -1.3 & -1.1 & -2.4 & -2.0 & -3.0 & \textcolor{mygreen}{1.7} & -3.2 & -2.3 \\
\midrule
\multirow{2}{*}{\ssc{Base}} & \prompt & 29.8$_{\smallsup{0.4}}$& 24.2$_{\smallsup{0.7}}$& 25.0$_{\smallsup{0.5}}$& 23.8$_{\smallsup{0.1}}$& 19.2$_{\smallsup{0.5}}$& 14.3$_{\smallsup{0.6}}$& 20.2$_{\smallsup{0.1}}$& 22.1$_{\smallsup{0.7}}$& 20.4$_{\smallsup{0.7}}$& 18.5$_{\smallsup{0.7}}$& 13.1$_{\smallsup{0.8}}$& 24.4$_{\smallsup{0.6}}$& 17.3$_{\smallsup{0.6}}$& 14.2$_{\smallsup{0.3}}$& 16.2$_{\smallsup{0.4}}$& 10.6$_{\smallsup{0.7}}$& 16.5$_{\smallsup{0.3}}$& 14.4$_{\smallsup{0.1}}$ \\
& \model & 39.6$_{\smallsup{0.4}}$& 23.3$_{\smallsup{0.1}}$& 23.8$_{\smallsup{0.8}}$& 22.4$_{\smallsup{0.3}}$& 18.8$_{\smallsup{0.2}}$& 15.3$_{\smallsup{0.2}}$& 20.3$_{\smallsup{0.2}}$& 23.0$_{\smallsup{0.2}}$& 20.1$_{\smallsup{0.2}}$& 17.4$_{\smallsup{0.2}}$& 15.1$_{\smallsup{0.5}}$& 21.9$_{\smallsup{0.3}}$& 17.9$_{\smallsup{0.2}}$& 14.6$_{\smallsup{0.3}}$& 17.3$_{\smallsup{0.1}}$& 9.1$_{\smallsup{0.2}}$& 17.8$_{\smallsup{0.1}}$& 17.5$_{\smallsup{0.1}}$ \\
& $\Delta_{\text{P-M}}$ & -9.8 & \textcolor{mygreen}{0.9} & \textcolor{mygreen}{1.2} & \textcolor{mygreen}{1.4} & \textcolor{mygreen}{0.4} & -1.0 & -0.1 & -0.9 & \textcolor{mygreen}{0.3} & \textcolor{mygreen}{1.1} & -2.0 & \textcolor{mygreen}{2.5} & -0.6 & -0.4 & -1.1 & \textcolor{mygreen}{1.5} & -1.3 & -3.1 \\
\midrule
\multirow{2}{*}{\ssc{Large}} & \prompt & 35.3$_{\smallsup{0.3}}$& 29.4$_{\smallsup{0.3}}$& 29.0$_{\smallsup{0.0}}$& 28.8$_{\smallsup{0.1}}$& 24.8$_{\smallsup{0.5}}$& 20.4$_{\smallsup{0.8}}$& 24.3$_{\smallsup{0.2}}$& 27.2$_{\smallsup{0.1}}$& 27.0$_{\smallsup{0.3}}$& 24.1$_{\smallsup{0.5}}$& 20.8$_{\smallsup{1.1}}$& 29.3$_{\smallsup{0.3}}$& 24.7$_{\smallsup{1.0}}$& 19.4$_{\smallsup{0.3}}$& 23.4$_{\smallsup{0.7}}$& 17.3$_{\smallsup{0.1}}$& 22.7$_{\smallsup{0.4}}$& 19.5$_{\smallsup{0.2}}$ \\
& \model & 42.6$_{\smallsup{0.2}}$& 29.7$_{\smallsup{0.1}}$& 30.3$_{\smallsup{0.3}}$& 27.8$_{\smallsup{0.6}}$& 23.5$_{\smallsup{0.8}}$& 17.4$_{\smallsup{1.0}}$& 25.6$_{\smallsup{0.7}}$& 26.9$_{\smallsup{0.7}}$& 25.3$_{\smallsup{0.5}}$& 23.7$_{\smallsup{1.7}}$& 19.2$_{\smallsup{0.6}}$& 27.2$_{\smallsup{0.8}}$& 25.9$_{\smallsup{0.7}}$& 22.1$_{\smallsup{0.7}}$& 23.9$_{\smallsup{0.7}}$& 12.7$_{\smallsup{0.4}}$& 22.1$_{\smallsup{0.4}}$& 20.6$_{\smallsup{0.6}}$ \\
& $\Delta_{\text{P-M}}$ & -7.3 & -0.3 & -1.3 & \textcolor{mygreen}{1.0} & \textcolor{mygreen}{1.3} & \textcolor{mygreen}{3.0} & -1.3 & \textcolor{mygreen}{0.3} & \textcolor{mygreen}{1.7} & \textcolor{mygreen}{0.4} & \textcolor{mygreen}{1.6} & \textcolor{mygreen}{2.1} & -1.2 & -2.7 & -0.5 & \textcolor{mygreen}{4.6} & \textcolor{mygreen}{0.6} & -1.1 \\
\midrule
\multirow{2}{*}{\ssc{XL}} & \prompt & 38.4$_{\smallsup{0.2}}$& 34.8$_{\smallsup{0.4}}$& 33.3$_{\smallsup{0.3}}$& 33.4$_{\smallsup{0.3}}$& 28.9$_{\smallsup{0.1}}$& 25.3$_{\smallsup{0.3}}$& 28.7$_{\smallsup{0.3}}$& 33.1$_{\smallsup{0.1}}$& 32.3$_{\smallsup{0.2}}$& 30.4$_{\smallsup{0.4}}$& 24.4$_{\smallsup{2.0}}$& 34.1$_{\smallsup{0.4}}$& 33.2$_{\smallsup{0.4}}$& 23.1$_{\smallsup{2.3}}$& 27.4$_{\smallsup{1.3}}$& 17.3$_{\smallsup{2.3}}$& 26.8$_{\smallsup{0.4}}$& 23.5$_{\smallsup{0.2}}$ \\
& \model & 45.0$_{\smallsup{0.3}}$& 32.2$_{\smallsup{0.3}}$& 33.1$_{\smallsup{0.3}}$& 31.9$_{\smallsup{0.5}}$& 25.3$_{\smallsup{1.0}}$& 19.7$_{\smallsup{0.6}}$& 28.6$_{\smallsup{0.2}}$& 28.3$_{\smallsup{0.5}}$& 28.4$_{\smallsup{0.7}}$& 27.3$_{\smallsup{0.8}}$& 30.0$_{\smallsup{0.8}}$& 29.8$_{\smallsup{0.7}}$& 25.6$_{\smallsup{0.5}}$& 25.4$_{\smallsup{0.7}}$& 29.1$_{\smallsup{0.4}}$& 16.3$_{\smallsup{0.5}}$& 23.5$_{\smallsup{0.6}}$& 22.9$_{\smallsup{0.3}}$ \\
& $\Delta_{\text{P-M}}$ & -6.6 & \textcolor{mygreen}{2.6} & \textcolor{mygreen}{0.2} & \textcolor{mygreen}{1.5} & \textcolor{mygreen}{3.6} & \textcolor{mygreen}{5.6} & \textcolor{mygreen}{0.1} & \textcolor{mygreen}{4.8} & \textcolor{mygreen}{3.9} & \textcolor{mygreen}{3.1} & -5.6 & \textcolor{mygreen}{4.3} & \textcolor{mygreen}{7.6} & -2.3 & -1.7 & \textcolor{mygreen}{1.0} & \textcolor{mygreen}{3.3} & \textcolor{mygreen}{0.6} \\
\midrule
\multirow{2}{*}{\ssc{XXL}} & \prompt & 43.4$_{\smallsup{0.4}}$& 36.8$_{\smallsup{0.4}}$& 36.1$_{\smallsup{0.4}}$& 37.4$_{\smallsup{0.2}}$& 30.3$_{\smallsup{0.4}}$& 29.2$_{\smallsup{1.0}}$& 30.9$_{\smallsup{0.5}}$& 35.1$_{\smallsup{0.6}}$& 35.1$_{\smallsup{0.5}}$& 32.9$_{\smallsup{0.3}}$& 31.9$_{\smallsup{3.2}}$& 38.0$_{\smallsup{0.0}}$& 37.4$_{\smallsup{0.7}}$& 27.0$_{\smallsup{1.6}}$& 33.6$_{\smallsup{0.9}}$& 17.9$_{\smallsup{5.1}}$& 30.7$_{\smallsup{0.1}}$& 25.8$_{\smallsup{0.9}}$ \\
& \model & 46.7$_{\smallsup{0.1}}$& 37.1$_{\smallsup{0.6}}$& 35.8$_{\smallsup{0.3}}$& 35.5$_{\smallsup{0.6}}$& 30.2$_{\smallsup{0.3}}$& 27.2$_{\smallsup{0.4}}$& 31.6$_{\smallsup{0.5}}$& 32.6$_{\smallsup{0.1}}$& 30.9$_{\smallsup{0.8}}$& 30.1$_{\smallsup{0.9}}$& 40.8$_{\smallsup{0.3}}$& 34.0$_{\smallsup{0.5}}$& 30.1$_{\smallsup{0.5}}$& 29.8$_{\smallsup{0.4}}$& 31.2$_{\smallsup{0.6}}$& 23.1$_{\smallsup{0.7}}$& 26.0$_{\smallsup{0.7}}$& 26.2$_{\smallsup{0.3}}$ \\
& $\Delta_{\text{P-M}}$ & -3.3 & -0.3 & \textcolor{mygreen}{0.3} & \textcolor{mygreen}{1.9} & \textcolor{mygreen}{0.1} & \textcolor{mygreen}{2.0} & -0.7 & \textcolor{mygreen}{2.5} & \textcolor{mygreen}{4.2} & \textcolor{mygreen}{2.8} & -8.9 & \textcolor{mygreen}{4.0} & \textcolor{mygreen}{7.3} & -2.8 & \textcolor{mygreen}{2.4} & -5.2 & \textcolor{mygreen}{4.7} & -0.4 \\
\bottomrule
\end{tabular}
\end{adjustbox}
\caption{Best validation \ssc{SP-Rouge} per language on \ssc{WikiLingua-0}.}
\label{tbl:wikilingua}
\end{table*}

\clearpage
\newpage
\section{Measuring the correlation between \bsprouge and human judgments}
\label{appendix:correlation}
\begin{figure}[t!]
\centering
\includegraphics[width=0.95\columnwidth]{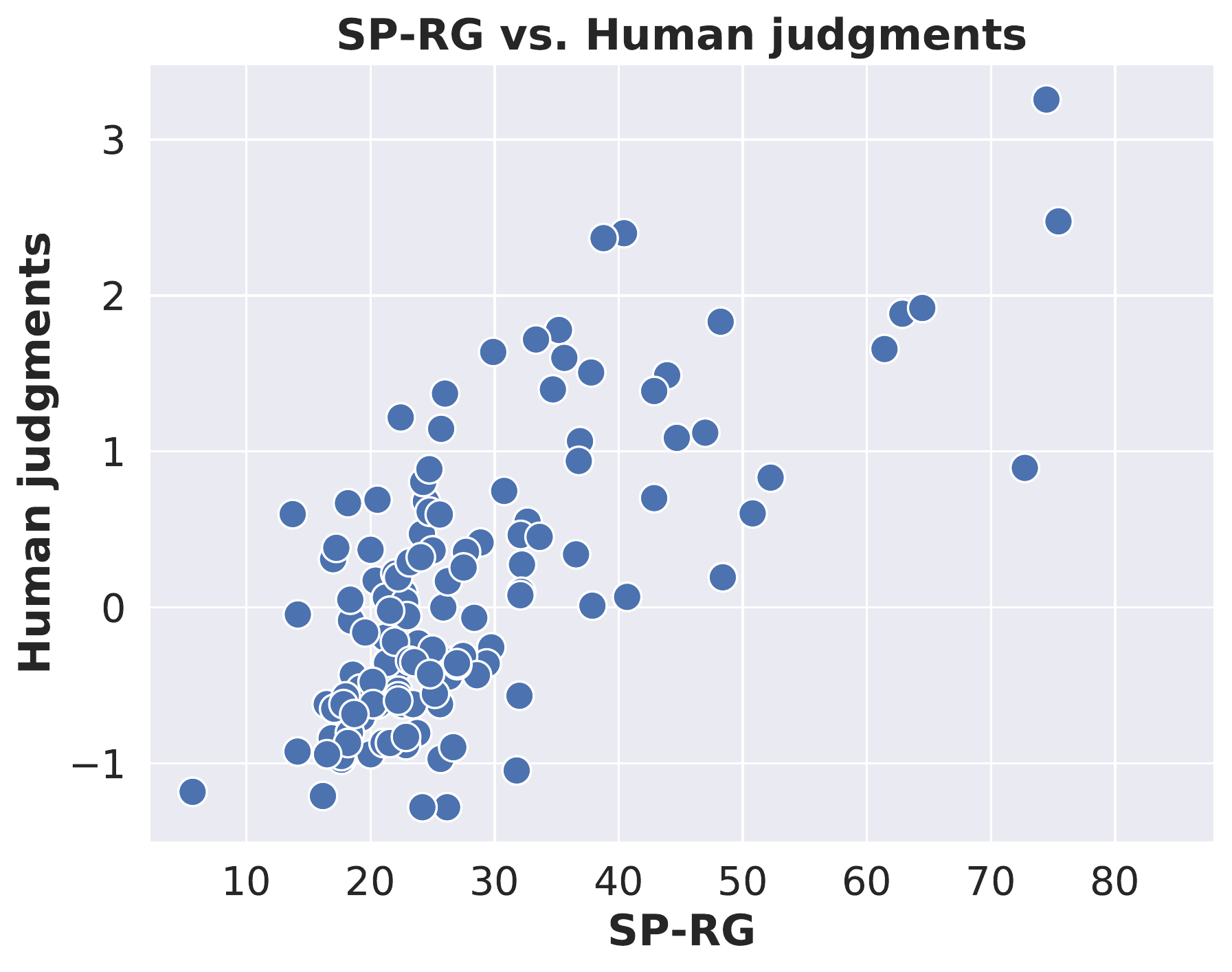}
\caption{A scatterplot demonstrating the linear relationship between \ssc{SP-Rouge} and human judgments on \ssc{Focus} for French summaries. As shown in Table~\ref{tbl:correlations}, \ssc{SP-Rouge} also correlates well with human judgments on other languages.}
\label{figure:correlation}
\vspace{-2mm}
\end{figure}

To evaluate how well our proposed \ssc{SP-Rouge} metric correlates with human judgments, we use the \ssc{MultiSumm Eval} dataset introduced by \citet{FKoto21}, which is a manually-annotated multilingual resource for summarization evaluation with $4{,}320$ human annotations on \ssc{Focus} (precision) and \ssc{Coverage} (recall) between machine-generated summaries and ground-truth summaries. We compare \ssc{SP-Rouge} to \ssc{BLEURT}~\cite{TSellam20}, which is a learned evaluation metric based on \ssc{BERT}~\cite{JDevlin19}. Table \ref{tbl:correlations} shows the Pearson correlation coefficient between these metrics and human judgments across 8 \ssc{MultiSumm Eval} languages, including German (\ssc{De}), English (\ssc{En}), Spanish (\ssc{Es}), French (\ssc{Fr}), Indonesian (\ssc{Id}), Russian (\ssc{Ru}), Turkish (\ssc{Tr}), and Mandarin Chinese (\ssc{Zh}). Overall, we found that the performance of \ssc{SP-Rouge} and the more computationally expensive \ssc{BLEURT} metric were similar. Specifically, \ssc{SP-Rouge} achieved an average \ssc{Focus} score of 0.68 and an average \ssc{Coverage} score of 0.65, whereas \ssc{BLEURT} achieved 0.68 and 0.70, respectively. Figure \ref{figure:correlation} demonstrates the linear relationship between \ssc{SP-Rouge-LSUM} vs \ssc{Focus} scores on French.
\begin{table*}[t!]
\centering
\begin{adjustbox}{max width=\textwidth}
\begin{tabular}{l@{\hspace{0.5\tabcolsep}}l ccccccccc ccccccccc}
\toprule
& \multicolumn{9}{c}{\bssc{Focus}} & \multicolumn{9}{c}{\bssc{Coverage}} \\
\cmidrule(lr){2-10}
\cmidrule(lr){11-19} 
\textbf{Metric} 
& \multicolumn{1}{c}{\ssc{De}} 
& \multicolumn{1}{c}{\ssc{En}} 
& \multicolumn{1}{c}{\ssc{Es}} 
& \multicolumn{1}{c}{\ssc{Fr}} 
& \multicolumn{1}{c}{\ssc{Id}} 
& \multicolumn{1}{c}{\ssc{Ru}} 
& \multicolumn{1}{c}{\ssc{Tr}} 
& \multicolumn{1}{c}{\ssc{Zh}} 
& \multicolumn{1}{c}{\ssc{Avg.}} 
& \multicolumn{1}{c}{\ssc{De}} 
& \multicolumn{1}{c}{\ssc{En}} 
& \multicolumn{1}{c}{\ssc{Es}} 
& \multicolumn{1}{c}{\ssc{Fr}} 
& \multicolumn{1}{c}{\ssc{Id}} 
& \multicolumn{1}{c}{\ssc{Ru}} 
& \multicolumn{1}{c}{\ssc{Tr}} 
& \multicolumn{1}{c}{\ssc{Zh}} 
& \multicolumn{1}{c}{\ssc{Avg.}}
\\
\midrule
\sprouge & 0.88 & 0.53 & 0.60 & 0.67  & 0.67 & 0.49 & 0.82 & 0.77 & 0.68 & 0.88 & 0.53 & 0.65 & 0.62  & 0.68 & 0.37 & 0.75 & 0.72 & 0.65 \\
\ssc{BLEURT} & 0.87 & 0.52 & 0.66 & 0.70 & 0.61 & 0.56 & 0.79 & 0.73 & 0.68 & 0.88 & 0.60 & 0.65 & 0.71 & 0.62 & 0.59 & 0.79 & 0.75 & 0.70 \\
\bottomrule
\end{tabular}
\end{adjustbox}
\caption{\ssc{SP-Rouge} correlates well with human judgments, providing a similar correlation to \ssc{BLEURT} while being significantly less computationally expensive.}%
\label{tbl:correlations}
\end{table*}

\begin{table*}[t!]
\centering
\begin{adjustbox}{max width=\textwidth}
\begin{tabular}{l@{\hspace{0.5\tabcolsep}}l ll lll lll lll lll}
\toprule
& & \multicolumn{2}{c}{\bssc{En}} & \multicolumn{3}{c}{\bssc{Fr}} & \multicolumn{3}{c}{\bssc{Ru}} & \multicolumn{3}{c}{\bssc{Vi}} & \multicolumn{3}{c}{\bssc{Th}} \\
\cmidrule(lr){3-4}
\cmidrule(lr){5-7} 
\cmidrule(lr){8-10}
\cmidrule(lr){11-13}
\cmidrule(lr){14-16}
\textbf{Size} & \textbf{Method} & \multicolumn{1}{c}{\fsprouge} & \multicolumn{1}{c}{\fssc{LID$_{\text{En}}$}} &
\multicolumn{1}{c}{\fsprouge} & \multicolumn{1}{c}{\fssc{LID$_{\text{En}}$}} &
\multicolumn{1}{c}{\fssc{LID$_{\text{Fr}}$}} &
\multicolumn{1}{c}{\fsprouge} & \multicolumn{1}{c}{\fssc{LID$_{\text{En}}$}} &
\multicolumn{1}{c}{\fssc{LID$_{\text{Ru}}$}} &
\multicolumn{1}{c}{\fsprouge} & \multicolumn{1}{c}{\fssc{LID$_{\text{En}}$}} &
\multicolumn{1}{c}{\fssc{LID$_{\text{Vi}}$}} &
\multicolumn{1}{c}{\fsprouge} & \multicolumn{1}{c}{\fssc{LID$_{\text{En}}$}} &
\multicolumn{1}{c}{\fssc{LID$_{\text{Th}}$}}
\\
\midrule
\midrule
\multirow{1}{*}{-} & \ssc{Lead-64} & 20.7$_{\smallsup{0.0}}$ & 99.6$_{\smallsup{0.0}}$ & 18.9$_{\smallsup{0.0}}$ & 0.0$_{\smallsup{0.0}}$ & 100.0$_{\smallsup{0.0}}$ & 16.5$_{\smallsup{0.0}}$ & 0.0$_{\smallsup{0.0}}$ & 99.6$_{\smallsup{0.0}}$ & 22.1$_{\smallsup{0.0}}$ & 0.0$_{\smallsup{0.0}}$ & 100.0$_{\smallsup{0.0}}$ & 15.9$_{\smallsup{0.0}}$ & 0.0$_{\smallsup{0.0}}$ & 97.6$_{\smallsup{0.0}}$ \\
\midrule
\multirow{1}{*}{\fssc{XXL}} & \prompt & 43.4$_{\smallsup{0.4}}$ & 92.0$_{\smallsup{0.5}}$ & 37.4$_{\smallsup{0.2}}$ & 2.9$_{\smallsup{1.5}}$ & 95.9$_{\smallsup{1.5}}$ & 29.2$_{\smallsup{1.0}}$ & 9.1$_{\smallsup{2.4}}$ & 84.4$_{\smallsup{1.8}}$ & 38.0$_{\smallsup{0.0}}$ & 1.8$_{\smallsup{1.1}}$ & 96.4$_{\smallsup{0.8}}$ & 37.4$_{\smallsup{0.7}}$ & 13.5$_{\smallsup{2.0}}$ & 75.5$_{\smallsup{1.5}}$\\
\multirow{1}{*}{\fssc{XXL}} & \prompt, \fssc{trans-test} & - & - & 37.0$_{\smallsup{0.4}}$ & 0.0$_{\smallsup{0.0}}$ & 98.9$_{\smallsup{0.2}}$ & 30.4$_{\smallsup{0.4}}$ & 0.0$_{\smallsup{0.0}}$ & 93.2$_{\smallsup{0.3}}$ & 37.5$_{\smallsup{0.1}}$ & 0.0$_{\smallsup{0.0}}$ & 99.9$_{\smallsup{0.1}}$ & 28.7$_{\smallsup{0.5}}$ & 0.0$_{\smallsup{0.0}}$ & 100.0$_{\smallsup{0.0}}$\\
\multirow{1}{*}{\fssc{XXL}} & \prompt, \fssc{trans-train} & - & - & 38.1$_{\smallsup{1.5}}$ & 0.0$_{\smallsup{0.0}}$ & 98.8$_{\smallsup{0.2}}$ & 31.3$_{\smallsup{0.2}}$ & 0.0$_{\smallsup{0.0}}$ & 94.3$_{\smallsup{0.8}}$ & 39.2$_{\smallsup{0.1}}$ & 0.0$_{\smallsup{0.0}}$ & 100.0$_{\smallsup{0.0}}$ & 37.1$_{\smallsup{0.3}}$ & 0.0$_{\smallsup{0.0}}$ & 100.0$_{\smallsup{0.0}}$\\
\multirow{1}{*}{\fssc{XXL}} & \prompt, \fssc{sup} & 43.4$_{\smallsup{0.4}}$ & 92.0$_{\smallsup{0.5}}$ & 41.0$_{\smallsup{0.1}}$ & 0.0$_{\smallsup{0.0}}$ & 99.3$_{\smallsup{0.1}}$ & 33.5$_{\smallsup{0.3}}$ & 0.0$_{\smallsup{0.0}}$ & 92.5$_{\smallsup{0.5}}$ &
38.8$_{\smallsup{0.3}}$ & 0.6$_{\smallsup{0.4}}$ & 96.7$_{\smallsup{0.9}}$ & 45.0$_{\smallsup{0.1}}$ & 0.1$_{\smallsup{0.1}}$ & 99.6$_{\smallsup{0.3}}$\\
\multirow{1}{*}{\fssc{XXL}} & \prompt, \fssc{sup-all} & 41.0$_{\smallsup{0.4}}$ & 90.4$_{\smallsup{0.7}}$ & 40.4$_{\smallsup{0.1}}$ & 0.2$_{\smallsup{0.3}}$ & 98.1$_{\smallsup{0.2}}$ & 33.3$_{\smallsup{0.2}}$ & 0.1$_{\smallsup{0.1}}$ & 91.4$_{\smallsup{1.6}}$ & 39.5$_{\smallsup{0.1}}$ & 0.4$_{\smallsup{0.3}}$ & 98.3$_{\smallsup{0.6}}$ & 44.8$_{\smallsup{0.7}}$ & 0.0$_{\smallsup{0.0}}$ & 100.0$_{\smallsup{0.0}}$ \\
\midrule
\multirow{1}{*}{\fssc{XXL}} & \model & 46.7$_{\smallsup{0.1}}$ & 94.4$_{\smallsup{0.8}}$ & 35.5$_{\smallsup{0.6}}$ & 9.1$_{\smallsup{3.1}}$ & 86.0$_{\smallsup{3.1}}$ & 27.2$_{\smallsup{0.4}}$ & 19.7$_{\smallsup{2.5}}$ & 57.5$_{\smallsup{2.8}}$ & 34.0$_{\smallsup{0.5}}$ & 14.8$_{\smallsup{3.5}}$ & 79.1$_{\smallsup{3.5}}$ & 30.1$_{\smallsup{0.5}}$ & 32.7$_{\smallsup{6.6}}$ & 16.8$_{\smallsup{3.6}}$\\
\multirow{1}{*}{\fssc{XXL}} & \model, \fssc{trans-test} & - & - & 38.9$_{\smallsup{0.1}}$ & 0.0$_{\smallsup{0.0}}$ & 98.9$_{\smallsup{0.1}}$ & 32.9$_{\smallsup{0.2}}$ & 0.0$_{\smallsup{0.0}}$ & 93.1$_{\smallsup{1.3}}$ & 39.2$_{\smallsup{0.4}}$ & 0.0$_{\smallsup{0.0}}$ & 99.5$_{\smallsup{0.4}}$ & 31.7$_{\smallsup{0.4}}$ & 0.0$_{\smallsup{0.0}}$ & 100.0$_{\smallsup{0.0}}$\\
\multirow{1}{*}{\fssc{XXL}} & \model, \fssc{trans-train} & - & - & 41.6$_{\smallsup{0.0}}$ & 0.4$_{\smallsup{0.0}}$ & 98.5$_{\smallsup{0.0}}$ & 34.9$_{\smallsup{0.1}}$ & 0.0$_{\smallsup{0.0}}$ & 95.4$_{\smallsup{0.6}}$ & 41.4$_{\smallsup{0.2}}$ & 0.0$_{\smallsup{0.0}}$ & 100.0$_{\smallsup{0.0}}$ & 38.7$_{\smallsup{0.5}}$ & 0.0$_{\smallsup{0.0}}$ & 100.0$_{\smallsup{0.0}}$\\
\multirow{1}{*}{\fssc{XXL}} & \model, \fssc{sup} & 46.7$_{\smallsup{0.1}}$ & 94.4$_{\smallsup{0.8}}$ & 43.8$_{\smallsup{0.2}}$ & 0.1$_{\smallsup{0.2}}$ & 99.2$_{\smallsup{0.6}}$ & 36.6$_{\smallsup{0.1}}$ & 0.0$_{\smallsup{0.0}}$ & 95.5$_{\smallsup{1.0}}$ & 
42.0$_{\smallsup{0.2}}$ & 0.0$_{\smallsup{0.0}}$ & 99.7$_{\smallsup{0.1}}$ & 48.8$_{\smallsup{0.5}}$ & 0.0$_{\smallsup{0.0}}$ & 99.9$_{\smallsup{0.2}}$ \\
\multirow{1}{*}{\fssc{XXL}} & \model, \fssc{sup-all} & 47.1$_{\smallsup{0.0}}$ & 93.8$_{\smallsup{0.8}}$ & 44.9$_{\smallsup{0.1}}$ & 0.0$_{\smallsup{0.0}}$ & 98.8$_{\smallsup{0.5}}$ & 37.6$_{\smallsup{0.2}}$ & 0.1$_{\smallsup{0.2}}$ & 93.7$_{\smallsup{1.0}}$ & 43.8$_{\smallsup{0.2}}$ & 0.0$_{\smallsup{0.0}}$ & 99.7$_{\smallsup{0.2}}$ & 50.2$_{\smallsup{0.1}}$ & 0.0$_{\smallsup{0.0}}$ & 100.0$_{\smallsup{0.0}}$\\
\midrule
\midrule
\multirow{1}{*}{\fssc{Small}} & \prompt & 24.5$_{\smallsup{0.2}}$ & 82.8$_{\smallsup{0.9}}$ & 19.2$_{\smallsup{0.1}}$ & 3.3$_{\smallsup{0.7}}$ & 77.4$_{\smallsup{2.7}}$ & 11.4$_{\smallsup{0.1}}$ & 29.6$_{\smallsup{1.7}}$ & 10.1$_{\smallsup{1.0}}$ & 21.4$_{\smallsup{0.2}}$ & 2.3$_{\smallsup{0.7}}$ & 87.2$_{\smallsup{2.4}}$ & 14.9$_{\smallsup{0.4}}$ & 45.9$_{\smallsup{2.6}}$ & 3.3$_{\smallsup{0.4}}$\\
\multirow{1}{*}{\fssc{Base}} & \prompt & 29.8$_{\smallsup{0.4}}$ & 85.2$_{\smallsup{0.9}}$ & 23.8$_{\smallsup{0.1}}$ & 5.6$_{\smallsup{2.9}}$ & 82.8$_{\smallsup{2.9}}$ & 14.3$_{\smallsup{0.6}}$ & 39.2$_{\smallsup{3.2}}$ & 24.5$_{\smallsup{5.9}}$ & 24.4$_{\smallsup{0.6}}$ & 6.0$_{\smallsup{1.4}}$ & 81.9$_{\smallsup{2.4}}$ & 17.3$_{\smallsup{0.6}}$ & 34.3$_{\smallsup{1.5}}$ & 33.5$_{\smallsup{2.5}}$\\
\multirow{1}{*}{\fssc{Large}} & \prompt & 35.3$_{\smallsup{0.3}}$ & 89.4$_{\smallsup{0.7}}$ & 28.8$_{\smallsup{0.1}}$ & 3.6$_{\smallsup{0.9}}$ & 91.1$_{\smallsup{0.8}}$ & 20.4$_{\smallsup{0.8}}$ & 13.3$_{\smallsup{2.6}}$ & 74.6$_{\smallsup{3.8}}$ & 29.3$_{\smallsup{0.3}}$ & 3.0$_{\smallsup{0.5}}$ & 89.3$_{\smallsup{2.0}}$ & 24.7$_{\smallsup{1.0}}$ & 29.0$_{\smallsup{7.6}}$ & 45.9$_{\smallsup{9.3}}$\\
\multirow{1}{*}{\fssc{XL}} & \prompt & 38.4$_{\smallsup{0.2}}$ & 90.5$_{\smallsup{0.4}}$ & 33.4$_{\smallsup{0.3}}$ & 2.4$_{\smallsup{0.8}}$ & 94.8$_{\smallsup{0.5}}$ & 25.3$_{\smallsup{0.3}}$ & 9.6$_{\smallsup{1.5}}$ & 79.3$_{\smallsup{1.6}}$ & 34.1$_{\smallsup{0.4}}$ & 3.4$_{\smallsup{0.3}}$ & 91.9$_{\smallsup{0.5}}$ & 33.2$_{\smallsup{0.4}}$ & 19.8$_{\smallsup{5.5}}$ & 66.0$_{\smallsup{6.8}}$\\
\multirow{1}{*}{\fssc{XXL}} & \prompt & 43.4$_{\smallsup{0.4}}$ & 92.0$_{\smallsup{0.5}}$ & \textbf{37.4}$_{\smallsup{0.2}}$ & 2.9$_{\smallsup{1.5}}$ & 95.9$_{\smallsup{1.5}}$ & \textbf{29.2}$_{\smallsup{1.0}}$ & 9.1$_{\smallsup{2.4}}$ & 84.4$_{\smallsup{1.8}}$ & \textbf{38.0}$_{\smallsup{0.0}}$ & 1.8$_{\smallsup{1.1}}$ & 96.4$_{\smallsup{0.8}}$ & \textbf{37.4}$_{\smallsup{0.7}}$ & 13.5$_{\smallsup{2.0}}$ & 75.5$_{\smallsup{1.5}}$\\
\midrule
\multirow{1}{*}{\fssc{Small}} & \model & 38.0$_{\smallsup{0.4}}$ & 92.4$_{\smallsup{0.4}}$ & 21.3$_{\smallsup{0.3}}$ & 72.4$_{\smallsup{2.2}}$ & 11.9$_{\smallsup{1.6}}$ & 14.6$_{\smallsup{0.2}}$ & 82.5$_{\smallsup{1.0}}$ & 0.0$_{\smallsup{0.1}}$ & 22.5$_{\smallsup{0.1}}$ & 39.9$_{\smallsup{4.8}}$ & 34.9$_{\smallsup{2.9}}$ & 17.3$_{\smallsup{0.1}}$ & 78.1$_{\smallsup{4.2}}$ & 0.1$_{\smallsup{0.1}}$\\
\multirow{1}{*}{\fssc{Base}} & \model & 39.6$_{\smallsup{0.4}}$ & 92.0$_{\smallsup{1.0}}$ & 22.4$_{\smallsup{0.3}}$ & 51.0$_{\smallsup{10.2}}$ & 25.3$_{\smallsup{7.4}}$ & 15.3$_{\smallsup{0.2}}$ & 79.0$_{\smallsup{11.7}}$ & 0.7$_{\smallsup{1.1}}$ & 21.9$_{\smallsup{0.3}}$ & 41.0$_{\smallsup{10.5}}$ & 34.0$_{\smallsup{8.3}}$ & 17.9$_{\smallsup{0.2}}$ & 89.0$_{\smallsup{0.8}}$ & 0.3$_{\smallsup{0.2}}$\\
\multirow{1}{*}{\fssc{Large}} & \model & 42.6$_{\smallsup{0.2}}$ & 92.8$_{\smallsup{0.3}}$ & 27.8$_{\smallsup{0.6}}$ & 9.9$_{\smallsup{4.1}}$ & 77.7$_{\smallsup{5.4}}$ & 17.4$_{\smallsup{1.0}}$ & 50.0$_{\smallsup{3.2}}$ & 21.4$_{\smallsup{3.8}}$ & 27.2$_{\smallsup{0.8}}$ & 13.6$_{\smallsup{6.0}}$ & 69.2$_{\smallsup{7.6}}$ & 25.9$_{\smallsup{0.7}}$ & 36.5$_{\smallsup{4.6}}$ & 35.4$_{\smallsup{2.1}}$\\
\multirow{1}{*}{\fssc{XL}} & \model & 45.0$_{\smallsup{0.3}}$ & 94.2$_{\smallsup{1.6}}$ & 31.9$_{\smallsup{0.5}}$ & 15.7$_{\smallsup{2.6}}$ & 76.2$_{\smallsup{3.9}}$ & 19.7$_{\smallsup{0.6}}$ & 61.6$_{\smallsup{15.8}}$ & 19.3$_{\smallsup{13.1}}$ & 29.8$_{\smallsup{0.7}}$ & 21.6$_{\smallsup{3.5}}$ & 64.8$_{\smallsup{4.5}}$ & 25.6$_{\smallsup{0.5}}$ & 54.7$_{\smallsup{14.5}}$ & 24.9$_{\smallsup{13.7}}$\\
\multirow{1}{*}{\fssc{XXL}} & \model & 46.7$_{\smallsup{0.1}}$ & 94.4$_{\smallsup{0.8}}$ & \textbf{35.5}$_{\smallsup{0.6}}$ & 9.1$_{\smallsup{3.1}}$ & 86.0$_{\smallsup{3.1}}$ & \textbf{27.2}$_{\smallsup{0.4}}$ & 19.7$_{\smallsup{2.5}}$ & 57.5$_{\smallsup{2.8}}$ & \textbf{34.0}$_{\smallsup{0.5}}$ & 14.8$_{\smallsup{3.5}}$ & 79.1$_{\smallsup{3.5}}$ & \textbf{30.1}$_{\smallsup{0.5}}$ & 32.7$_{\smallsup{6.6}}$ & 16.8$_{\smallsup{3.6}}$\\
\midrule
\midrule
\multirow{4}{*}{\fssc{Base}} & \prompt, \fssc{l=1} & 19.7$_{\smallsup{0.1}}$ & 75.9$_{\smallsup{0.8}}$ & 18.0$_{\smallsup{0.1}}$ & 0.9$_{\smallsup{0.2}}$ & 89.0$_{\smallsup{0.2}}$ & 14.8$_{\smallsup{0.1}}$ & 2.1$_{\smallsup{0.3}}$ & 83.4$_{\smallsup{0.2}}$ & 19.1$_{\smallsup{0.1}}$ & 0.2$_{\smallsup{0.0}}$ & 92.4$_{\smallsup{0.5}}$ & 19.2$_{\smallsup{0.1}}$ & 3.3$_{\smallsup{2.4}}$ & 80.2$_{\smallsup{12.2}}$\\
  & \prompt, \fssc{l=10} & 25.1$_{\smallsup{0.1}}$ & 84.4$_{\smallsup{1.2}}$ & 21.6$_{\smallsup{0.2}}$ & 0.3$_{\smallsup{0.1}}$ & 91.7$_{\smallsup{1.0}}$ & \textbf{17.2}$_{\smallsup{0.5}}$ & 6.6$_{\smallsup{3.1}}$ & 76.4$_{\smallsup{6.4}}$ & 23.5$_{\smallsup{0.1}}$ & 0.5$_{\smallsup{0.2}}$ & 94.8$_{\smallsup{2.1}}$ & \textbf{21.0}$_{\smallsup{0.5}}$ & 11.8$_{\smallsup{0.8}}$ & 53.7$_{\smallsup{2.1}}$\\
  & \prompt, \fssc{l=100} & 29.8$_{\smallsup{0.4}}$ & 85.2$_{\smallsup{0.9}}$ & \textbf{23.8}$_{\smallsup{0.1}}$ & 5.6$_{\smallsup{2.9}}$ & 82.8$_{\smallsup{2.9}}$ & 14.3$_{\smallsup{0.6}}$ & 39.2$_{\smallsup{3.2}}$ & 24.5$_{\smallsup{5.9}}$ & \textbf{24.4}$_{\smallsup{0.6}}$ & 6.0$_{\smallsup{1.4}}$ & 81.9$_{\smallsup{2.4}}$ & 17.3$_{\smallsup{0.6}}$ & 34.3$_{\smallsup{1.5}}$ & 33.5$_{\smallsup{2.5}}$\\
  & \prompt, \fssc{l=1000} & 32.4$_{\smallsup{0.3}}$ & 86.2$_{\smallsup{1.1}}$ & 22.0$_{\smallsup{0.9}}$ & 8.8$_{\smallsup{2.0}}$ & 77.1$_{\smallsup{4.3}}$ & 14.0$_{\smallsup{0.5}}$ & 41.9$_{\smallsup{4.6}}$ & 19.5$_{\smallsup{3.9}}$ & 23.3$_{\smallsup{0.5}}$ & 8.4$_{\smallsup{0.8}}$ & 79.4$_{\smallsup{1.4}}$ & 16.3$_{\smallsup{1.0}}$ & 47.5$_{\smallsup{3.7}}$ & 18.9$_{\smallsup{4.9}}$\\
\midrule
\multirow{4}{*}{\fssc{XXL}} & \prompt, \fssc{l=1} & 37.8$_{\smallsup{0.1}}$ & 88.8$_{\smallsup{0.6}}$ & 35.0$_{\smallsup{0.3}}$ & 0.0$_{\smallsup{0.0}}$ & 99.2$_{\smallsup{0.2}}$ & 29.8$_{\smallsup{0.2}}$ & 0.3$_{\smallsup{0.2}}$ & 93.7$_{\smallsup{0.5}}$ & 36.3$_{\smallsup{0.2}}$ & 0.0$_{\smallsup{0.0}}$ & 98.7$_{\smallsup{0.3}}$ & 36.4$_{\smallsup{1.7}}$ & 0.1$_{\smallsup{0.1}}$ & 99.3$_{\smallsup{0.2}}$\\
  & \prompt, \fssc{l=10} & 41.2$_{\smallsup{0.4}}$ & 89.8$_{\smallsup{1.0}}$ & \textbf{37.6}$_{\smallsup{0.3}}$ & 0.0$_{\smallsup{0.0}}$ & 99.2$_{\smallsup{0.5}}$ & \textbf{31.3}$_{\smallsup{0.1}}$ & 1.0$_{\smallsup{0.1}}$ & 92.7$_{\smallsup{1.1}}$ & \textbf{38.3}$_{\smallsup{0.1}}$ & 0.0$_{\smallsup{0.0}}$ & 99.5$_{\smallsup{0.2}}$ & \textbf{41.2}$_{\smallsup{0.2}}$ & 2.0$_{\smallsup{1.2}}$ & 91.3$_{\smallsup{1.3}}$\\
  & \prompt, \fssc{l=100} & 43.4$_{\smallsup{0.4}}$ & 92.0$_{\smallsup{0.5}}$ & 37.4$_{\smallsup{0.2}}$ & 2.9$_{\smallsup{1.5}}$ & 95.9$_{\smallsup{1.5}}$ & 29.2$_{\smallsup{1.0}}$ & 9.1$_{\smallsup{2.4}}$ & 84.4$_{\smallsup{1.8}}$ & 38.0$_{\smallsup{0.0}}$ & 1.8$_{\smallsup{1.1}}$ & 96.4$_{\smallsup{0.8}}$ & 37.4$_{\smallsup{0.7}}$ & 13.5$_{\smallsup{2.0}}$ & 75.5$_{\smallsup{1.5}}$\\
  & \prompt, \fssc{l=1000} & 40.8$_{\smallsup{2.2}}$ & 92.0$_{\smallsup{2.0}}$ & 35.7$_{\smallsup{1.0}}$ & 1.5$_{\smallsup{0.5}}$ & 97.3$_{\smallsup{0.6}}$ & 28.8$_{\smallsup{0.4}}$ & 7.0$_{\smallsup{2.1}}$ & 85.9$_{\smallsup{2.7}}$ & 37.0$_{\smallsup{1.2}}$ & 0.8$_{\smallsup{0.6}}$ & 97.8$_{\smallsup{1.3}}$ & 37.8$_{\smallsup{1.2}}$ & 7.4$_{\smallsup{0.1}}$ & 81.7$_{\smallsup{3.4}}$\\
\bottomrule
\end{tabular}
\end{adjustbox}
\caption{Summarization quality (\ssc{SP-Rouge}) and language identification confidence scores (\ssc{LID}) across model sizes and methods (numbers in the subscript
indicate the standard deviation across 3 random seeds). Our results suggest that \ssc{WikiLingua-0} is a challenging task for both \modeltuning and \prompttuning.
As model size increases, \prompttuning usually produces better results than \modeltuning when there is a significant language shift at inference time.  
Longer prompts help to better learn the English summarization task.  However, the increased capacity leads the model to forgets other languages.
}
\label{tbl:challenge_wikilingua0}
\end{table*}
\section{Zero-shot evaluation results on \bssc{WikiLingua-0}}
\label{appendix:model_adaptation_wikilingua0}
Our zero-shot evaluation results on \ssc{WikiLingua-0} for French (\bssc{Fr}), Vietnamese (\bssc{Vi}), Russian (\bssc{Ru}), and Thai (\bssc{Th}) are shown in Table~\ref{tbl:challenge_wikilingua0}. See Table~\ref{tbl:wikilingua} for results across all target languages. Our results suggest that \ssc{WikiLingua-0} is a challenging task for both \modeltuning and \prompttuning.
As model size increases, \prompttuning usually produces better results than \modeltuning when there is a significant language shift at inference time.  
Longer prompts help to better learn the English summarization task.  However, the increased capacity leads the model to forgets other languages.

\section{Language-Specific Prompt Clustering Analysis}
\label{appendix:clustering}

\begin{figure*}
\begin{center}
\includegraphics[width=\textwidth]{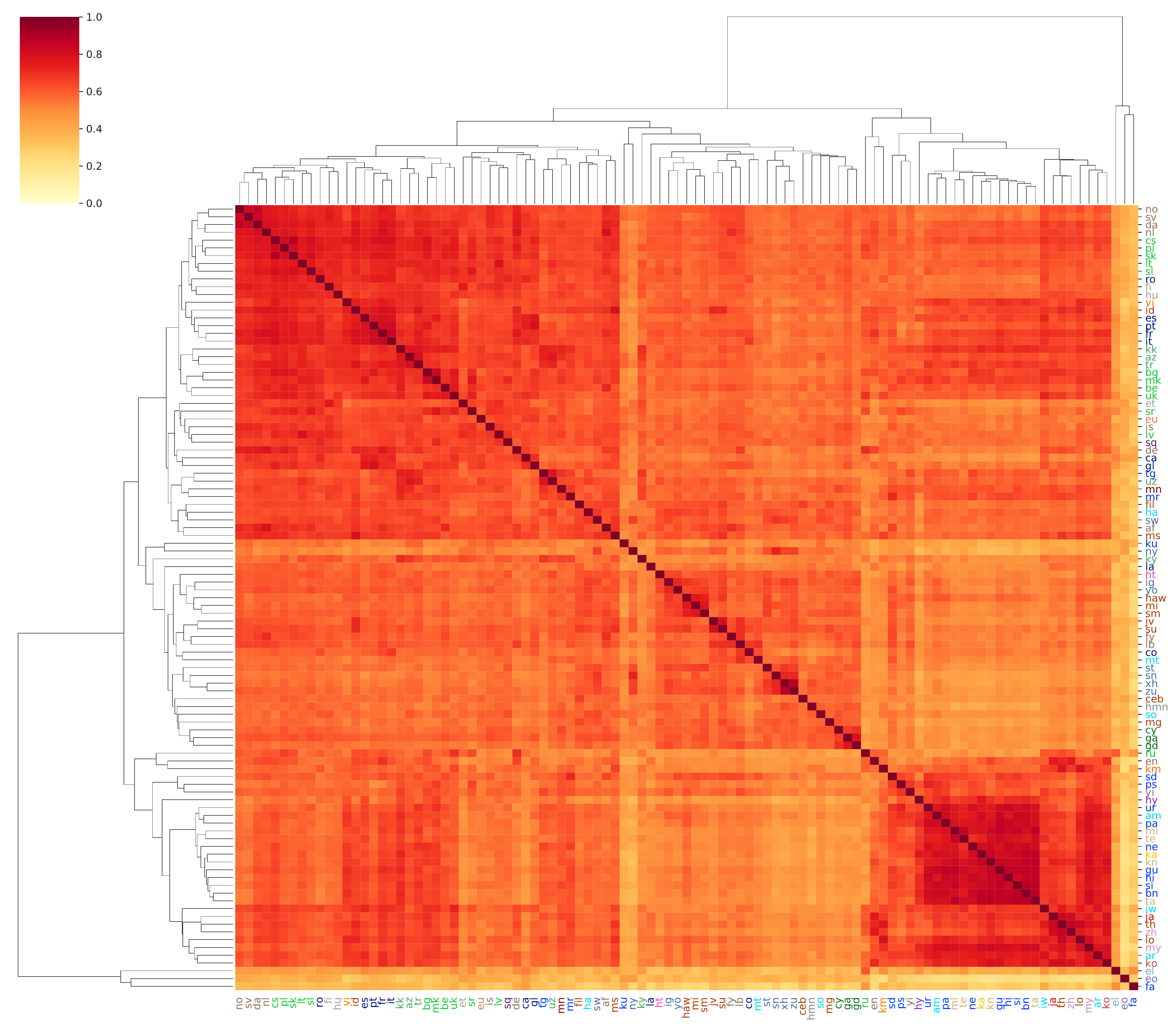}
\caption{A clustered heatmap of cosine similarities between 107 mT5-\ssc{Base} prompts trained on language-specific \ssc{LM} tasks. Language codes with the same color share a linguistic family.}
\label{fig:clustering}
\end{center}
\end{figure*}

To confirm that language-specific prompts trained on an LM task encode meaningful differences between languages, we train 107 prompts, one for each language in the \mcfour corpus. Specifically, we train prompts for the mT5-\ssc{Base} model, with a prompt length of 1, for 10K training steps, using a batch size of 32. The training task consists of classic causal language modeling, with an empty string fed as inputs to the encoder, and the document text passed as targets. Each prompt is trained exclusively on data from a single language bucket; however, we note that \mcfour contains a non-trivial number of language ID errors, particularly for lower-resource languages~\cite{JKreutzer22}.

Figure \ref{fig:clustering} shows a clustered heatmap of the cosine similarities between the trained prompts. We observe a number of interpretable clusters that give us confidence that the learned prompts encode meaningful language representations. For example, the leftmost 25 languages form a visible cluster and are all nearly all languages of Europe,\footnote{The only exceptions are Vietnamese (\ssc{vi}) and Indonesian (\ssc{id}), which are both written with Latin(-derived) scripts. We also note that Indonesian has a high language ID error rate within \mcfour.} with meaningful sub-clusters for different European regions: Northern (\ssc{No}, \ssc{Sv}, \ssc{Da}, \ssc{Nl}), Central (\ssc{Cs}, \ssc{Pl}, \ssc{Sk}, \ssc{Lt}, \ssc{Sl}), South-Western (\ssc{Es}, \ssc{Pt}, \ssc{Fr}, \ssc{It}) and Eastern (\ssc{Kk}, \ssc{Az}, \ssc{Tr}, \ssc{Bg}, \ssc{Mk}, \ssc{Be}, \ssc{Uk}). Another prominent cluster covers languages of India, Pakistan and Nepal (\ssc{Ml}, \ssc{Te}, \ssc{Ne}, \ssc{Ka}, \ssc{Kn}, \ssc{Gu}, \ssc{Hi}, \ssc{Si}, \ssc{Bn}, \ssc{Ta}), despite the fact that these languages cover different linguistic families and are written with different scripts. While geography seems to be the primary factor influencing prompt similarity, linguistic relationships also play a role. For instance, we observe that Finnish (\ssc{Fi}) and Hungarian (\ssc{Hu}), both Finno-Ugric languages, form a cluster despite their geographic distance. Similarly, Igbo (\ssc{Ig}), spoken mainly in Nigeria, is clustered nearby Haitian Creole (\ssc{Ht}), whose grammar derives from Igbo.

\section{Mitigating catastrophic forgetting}
\label{appendix:our_methods}
\begin{table*}[t!]
\centering
\begin{adjustbox}{max width=\textwidth}
\begin{tabular}{l@{\hspace{0.5\tabcolsep}}l ll lll lll lll lll}
\toprule
& & \multicolumn{2}{c}{\bssc{En}} & \multicolumn{3}{c}{\bssc{Fr}} & \multicolumn{3}{c}{\bssc{Ru}} & \multicolumn{3}{c}{\bssc{Vi}} & \multicolumn{3}{c}{\bssc{Th}} \\
\cmidrule(lr){3-4}
\cmidrule(lr){5-7} 
\cmidrule(lr){8-10}
\cmidrule(lr){11-13}
\cmidrule(lr){14-16}
\textbf{Size} & \textbf{Method} & \multicolumn{1}{c}{\fsprouge} & \multicolumn{1}{c}{\fssc{LID$_{\text{En}}$}} &
\multicolumn{1}{c}{\fsprouge} & \multicolumn{1}{c}{\fssc{LID$_{\text{En}}$}} &
\multicolumn{1}{c}{\fssc{LID$_{\text{Fr}}$}} &
\multicolumn{1}{c}{\fsprouge} & \multicolumn{1}{c}{\fssc{LID$_{\text{En}}$}} &
\multicolumn{1}{c}{\fssc{LID$_{\text{Ru}}$}} &
\multicolumn{1}{c}{\fsprouge} & \multicolumn{1}{c}{\fssc{LID$_{\text{En}}$}} &
\multicolumn{1}{c}{\fssc{LID$_{\text{Vi}}$}} &
\multicolumn{1}{c}{\fsprouge} & \multicolumn{1}{c}{\fssc{LID$_{\text{En}}$}} &
\multicolumn{1}{c}{\fssc{LID$_{\text{Th}}$}}
\\
\midrule
\midrule
\multirow{1}{*}{-} & \ssc{Lead-64} & 20.7$_{\smallsup{0.0}}$ & 99.6$_{\smallsup{0.0}}$ & 18.9$_{\smallsup{0.0}}$ & 0.0$_{\smallsup{0.0}}$ & 100.0$_{\smallsup{0.0}}$ & 16.5$_{\smallsup{0.0}}$ & 0.0$_{\smallsup{0.0}}$ & 99.6$_{\smallsup{0.0}}$ & 22.1$_{\smallsup{0.0}}$ & 0.0$_{\smallsup{0.0}}$ & 100.0$_{\smallsup{0.0}}$ & 15.9$_{\smallsup{0.0}}$ & 0.0$_{\smallsup{0.0}}$ & 97.6$_{\smallsup{0.0}}$ \\
\midrule
\midrule
\multirow{1}{*}{\fssc{Base}} & \prompt & 29.8$_{\smallsup{0.4}}$ & 85.2$_{\smallsup{0.9}}$ & 23.8$_{\smallsup{0.1}}$ & 5.6$_{\smallsup{2.9}}$ & 82.8$_{\smallsup{2.9}}$ & 14.3$_{\smallsup{0.6}}$ & 39.2$_{\smallsup{3.2}}$ & 24.5$_{\smallsup{5.9}}$ & 24.4$_{\smallsup{0.6}}$ & 6.0$_{\smallsup{1.4}}$ & 81.9$_{\smallsup{2.4}}$ & 17.3$_{\smallsup{0.6}}$ & 34.3$_{\smallsup{1.5}}$ & 33.5$_{\smallsup{2.5}}$\\
\multirow{1}{*}{\fssc{Base}} & \prompt, \fssc{Mix-Unsup} & 23.5$_{\smallsup{0.1}}$ & 83.4$_{\smallsup{1.4}}$ &
20.3$_{\smallsup{0.8}}$ & 0.2$_{\smallsup{0.3}}$ & 95.5$_{\smallsup{2.3}}$ & 16.1$_{\smallsup{0.3}}$ & 6.7$_{\smallsup{4.0}}$ & 77.5$_{\smallsup{8.2}}$ & 23.1$_{\smallsup{0.3}}$ & 0.3$_{\smallsup{0.2}}$ & 96.6$_{\smallsup{1.0}}$ & 20.9$_{\smallsup{0.8}}$ & 4.1$_{\smallsup{2.1}}$ & 76.9$_{\smallsup{7.0}}$ \\
\multirow{1}{*}{\fssc{Base}} & \prompt, \fssc{Mix-Unsup-All}& 23.0$_{\smallsup{0.4}}$ & 81.1$_{\smallsup{1.6}}$ & 19.3$_{\smallsup{1.0}}$ & 0.2$_{\smallsup{0.2}}$ & 92.0$_{\smallsup{2.2}}$ & 16.5$_{\smallsup{1.0}}$ & 2.1$_{\smallsup{1.1}}$ & 87.1$_{\smallsup{1.5}}$ & 22.7$_{\smallsup{0.8}}$ & 0.8$_{\smallsup{0.8}}$ & 96.1$_{\smallsup{1.5}}$ & 21.4$_{\smallsup{0.7}}$ & 2.8$_{\smallsup{1.1}}$ & 84.5$_{\smallsup{5.7}}$\\
\multirow{1}{*}{\fssc{Base}} & \prompt, \fssc{IT-Gigaword} & 30.8$_{\smallsup{0.2}}$ & 86.0$_{\smallsup{0.5}}$ & 24.0$_{\smallsup{0.2}}$ & 3.1$_{\smallsup{1.6}}$ & 85.5$_{\smallsup{0.7}}$ & 15.1$_{\smallsup{0.6}}$ & 41.7$_{\smallsup{5.5}}$ & 25.8$_{\smallsup{7.9}}$ & 24.8$_{\smallsup{0.0}}$ & 6.5$_{\smallsup{1.2}}$ & 81.4$_{\smallsup{0.9}}$ & 19.3$_{\smallsup{0.3}}$ & 33.5$_{\smallsup{3.1}}$ & 28.4$_{\smallsup{4.0}}$\\
\multirow{1}{*}{\fssc{Base}} & \prompt, \fssc{IT-LM} & 30.3$_{\smallsup{0.2}}$ & 86.2$_{\smallsup{0.2}}$ & \textbf{24.2}$_{\smallsup{0.1}}$ & 5.4$_{\smallsup{2.0}}$ & 83.0$_{\smallsup{2.3}}$ & 15.7$_{\smallsup{0.5}}$ & 36.0$_{\smallsup{2.1}}$ & 34.4$_{\smallsup{5.2}}$ & 24.3$_{\smallsup{0.2}}$ & 6.2$_{\smallsup{1.8}}$ & 81.0$_{\smallsup{1.4}}$ & 17.8$_{\smallsup{1.4}}$ & 41.2$_{\smallsup{6.6}}$ & 24.1$_{\smallsup{7.7}}$ \\
\multirow{1}{*}{\fssc{Base}} & \prompt, \fssc{FP-En} & 28.9$_{\smallsup{0.2}}$ & 84.7$_{\smallsup{0.3}}$ & 23.2$_{\smallsup{0.4}}$ & 3.4$_{\smallsup{0.9}}$ & 86.4$_{\smallsup{1.7}}$ & 16.1$_{\smallsup{0.6}}$ & 26.4$_{\smallsup{3.3}}$ & 48.5$_{\smallsup{4.2}}$ & \textbf{24.8}$_{\smallsup{0.7}}$ & 4.3$_{\smallsup{1.3}}$ & 84.6$_{\smallsup{3.1}}$ & 19.4$_{\smallsup{0.4}}$ & 28.6$_{\smallsup{4.4}}$ & 32.4$_{\smallsup{4.0}}$\\
\multirow{1}{*}{\fssc{Base}} & \prompt, \fssc{FP} & 28.9$_{\smallsup{0.2}}$ & 84.7$_{\smallsup{0.3}}$ & 23.6$_{\smallsup{0.4}}$ & 1.2$_{\smallsup{0.7}}$ & 93.0$_{\smallsup{1.3}}$ & \textbf{17.8}$_{\smallsup{0.8}}$ & 15.3$_{\smallsup{2.1}}$ & 64.5$_{\smallsup{1.1}}$ & 24.7$_{\smallsup{0.5}}$ & 2.1$_{\smallsup{0.8}}$ & 90.0$_{\smallsup{2.2}}$ & \textbf{21.1}$_{\smallsup{0.8}}$ & 19.8$_{\smallsup{5.1}}$ & 40.0$_{\smallsup{13.5}}$ \\
\midrule
\multirow{1}{*}{\fssc{Base}} & \model & 39.6$_{\smallsup{0.4}}$ & 92.0$_{\smallsup{1.0}}$ & 22.4$_{\smallsup{0.3}}$ & 51.0$_{\smallsup{10.2}}$ & 25.3$_{\smallsup{7.4}}$ & 15.3$_{\smallsup{0.2}}$ & 79.0$_{\smallsup{11.7}}$ & 0.7$_{\smallsup{1.1}}$ & 21.9$_{\smallsup{0.3}}$ & 41.0$_{\smallsup{10.5}}$ & 34.0$_{\smallsup{8.3}}$ & 17.9$_{\smallsup{0.2}}$ & 89.0$_{\smallsup{0.8}}$ & 0.3$_{\smallsup{0.2}}$\\
\multirow{1}{*}{\fssc{Base}} & \model, \fssc{Mix-Unsup} & 39.9$_{\smallsup{0.8}}$ & 93.6$_{\smallsup{1.4}}$ & \textbf{30.0}$_{\smallsup{0.5}}$ & 2.6$_{\smallsup{0.6}}$ & 90.5$_{\smallsup{1.1}}$ & \textbf{24.1}$_{\smallsup{0.8}}$ & 6.6$_{\smallsup{0.9}}$ & 73.5$_{\smallsup{4.2}}$ & \textbf{31.1}$_{\smallsup{0.2}}$ & 3.2$_{\smallsup{0.1}}$ & 90.4$_{\smallsup{0.7}}$ & 25.2$_{\smallsup{0.4}}$ & 16.2$_{\smallsup{2.9}}$ & 56.8$_{\smallsup{0.6}}$ \\
\multirow{1}{*}{\fssc{Base}} & \model, \fssc{Mix-Unsup-All} & 39.7$_{\smallsup{0.3}}$ & 93.0$_{\smallsup{1.3}}$ & 29.3$_{\smallsup{0.1}}$ & 5.5$_{\smallsup{1.0}}$ & 85.5$_{\smallsup{0.5}}$ & 21.7$_{\smallsup{0.4}}$ & 26.9$_{\smallsup{4.6}}$ & 41.8$_{\smallsup{2.2}}$ & 29.6$_{\smallsup{0.5}}$ & 8.9$_{\smallsup{1.3}}$ & 78.5$_{\smallsup{1.6}}$ & \textbf{25.5}$_{\smallsup{0.3}}$ & 24.6$_{\smallsup{2.2}}$ & 43.9$_{\smallsup{7.2}}$\\
\multirow{1}{*}{\fssc{Base}} & \model, \fssc{IT-Gigaword} & 40.5$_{\smallsup{0.3}}$ & 93.0$_{\smallsup{0.7}}$ & 20.8$_{\smallsup{0.1}}$ & 86.0$_{\smallsup{4.4}}$ & 4.0$_{\smallsup{1.1}}$ & 15.5$_{\smallsup{0.1}}$ & 92.5$_{\smallsup{0.3}}$ & 0.0$_{\smallsup{0.0}}$ & 21.2$_{\smallsup{0.1}}$ & 81.1$_{\smallsup{3.5}}$ & 6.3$_{\smallsup{1.6}}$ & 17.3$_{\smallsup{0.1}}$ & 93.4$_{\smallsup{2.2}}$ & 0.0$_{\smallsup{0.0}}$\\
\multirow{1}{*}{\fssc{Base}} & \model, \fssc{IT-LM} & 40.9$_{\smallsup{0.2}}$ & 93.3$_{\smallsup{1.1}}$ & 18.7$_{\smallsup{0.8}}$ & 61.8$_{\smallsup{43.7}}$ & 9.9$_{\smallsup{11.6}}$ & 15.7$_{\smallsup{0.1}}$ & 90.7$_{\smallsup{1.8}}$ & 0.2$_{\smallsup{0.2}}$ & 21.3$_{\smallsup{0.2}}$ & 65.9$_{\smallsup{5.1}}$ & 14.4$_{\smallsup{3.6}}$ & 17.2$_{\smallsup{0.1}}$ & 92.4$_{\smallsup{1.9}}$ & 0.1$_{\smallsup{0.2}}$ \\
\midrule
\midrule
\multirow{1}{*}{\fssc{XXL}} & \prompt & 43.4$_{\smallsup{0.4}}$ & 92.0$_{\smallsup{0.5}}$ & \textbf{37.4}$_{\smallsup{0.2}}$ & 2.9$_{\smallsup{1.5}}$ & 95.9$_{\smallsup{1.5}}$ & 29.2$_{\smallsup{1.0}}$ & 9.1$_{\smallsup{2.4}}$ & 84.4$_{\smallsup{1.8}}$ & \textbf{38.0}$_{\smallsup{0.0}}$ & 1.8$_{\smallsup{1.1}}$ & 96.4$_{\smallsup{0.8}}$ & 37.4$_{\smallsup{0.7}}$ & 13.5$_{\smallsup{2.0}}$ & 75.5$_{\smallsup{1.5}}$\\
\multirow{1}{*}{\fssc{XXL}} & \prompt, \fssc{Mix-Unsup} & 41.9$_{\smallsup{0.2}}$ & 90.1$_{\smallsup{0.8}}$ & 36.9$_{\smallsup{1.1}}$ & 1.1$_{\smallsup{0.6}}$ & 96.8$_{\smallsup{0.9}}$ & 26.2$_{\smallsup{3.0}}$ & 14.5$_{\smallsup{10.1}}$ & 72.3$_{\smallsup{13.1}}$ & 37.2$_{\smallsup{0.8}}$ & 1.3$_{\smallsup{0.9}}$ & 96.0$_{\smallsup{2.1}}$ & 37.4$_{\smallsup{2.0}}$ & 16.2$_{\smallsup{9.9}}$ & 74.0$_{\smallsup{10.8}}$ \\
\multirow{1}{*}{\fssc{XXL}} & \prompt, \fssc{Mix-Unsup-All} & 41.2$_{\smallsup{1.6}}$ & 91.2$_{\smallsup{1.1}}$ & 37.2$_{\smallsup{0.9}}$ & 1.5$_{\smallsup{0.6}}$ & 97.2$_{\smallsup{0.4}}$ & \textbf{30.0}$_{\smallsup{0.4}}$ & 3.9$_{\smallsup{1.1}}$ & 89.7$_{\smallsup{1.5}}$ & 37.3$_{\smallsup{1.1}}$ & 1.8$_{\smallsup{0.8}}$ & 96.0$_{\smallsup{1.7}}$ & \textbf{38.2}$_{\smallsup{2.0}}$ & 9.7$_{\smallsup{6.6}}$ & 81.9$_{\smallsup{6.4}}$\\
\multirow{1}{*}{\fssc{XXL}} & \prompt, \fssc{IT-Gigaword} & 43.5$_{\smallsup{0.1}}$ & 92.6$_{\smallsup{0.2}}$ & 36.6$_{\smallsup{0.5}}$ & 3.9$_{\smallsup{1.1}}$ & 94.2$_{\smallsup{1.5}}$ & 24.0$_{\smallsup{1.1}}$ & 37.5$_{\smallsup{5.7}}$ & 54.6$_{\smallsup{6.8}}$ & 37.2$_{\smallsup{0.2}}$ & 5.1$_{\smallsup{1.4}}$ & 93.2$_{\smallsup{1.5}}$ & 32.2$_{\smallsup{1.7}}$ & 33.7$_{\smallsup{6.0}}$ & 52.8$_{\smallsup{7.0}}$\\
\multirow{1}{*}{\fssc{XXL}} & \prompt, \fssc{IT-LM} & 42.9$_{\smallsup{0.1}}$ & 92.8$_{\smallsup{2.2}}$ & 36.4$_{\smallsup{0.5}}$ & 6.6$_{\smallsup{1.2}}$ & 91.4$_{\smallsup{2.0}}$ & 26.9$_{\smallsup{1.8}}$ & 17.9$_{\smallsup{7.2}}$ & 73.1$_{\smallsup{7.8}}$ & 37.2$_{\smallsup{0.3}}$ & 2.2$_{\smallsup{0.7}}$ & 94.3$_{\smallsup{1.4}}$ & \textbf{38.2}$_{\smallsup{0.2}}$ & 6.5$_{\smallsup{0.4}}$ & 83.1$_{\smallsup{1.8}}$ \\
\multirow{1}{*}{\fssc{XXL}} & \prompt, \fssc{FP-En} & 40.8$_{\smallsup{2.6}}$ & 90.0$_{\smallsup{3.0}}$ & 36.5$_{\smallsup{1.2}}$ & 2.5$_{\smallsup{1.6}}$ & 95.5$_{\smallsup{1.9}}$ & 27.9$_{\smallsup{1.2}}$ & 9.4$_{\smallsup{7.8}}$ & 81.3$_{\smallsup{9.3}}$ & 37.5$_{\smallsup{1.3}}$ & 0.4$_{\smallsup{0.3}}$ & 98.0$_{\smallsup{0.8}}$ & 36.7$_{\smallsup{0.6}}$ & 10.8$_{\smallsup{6.1}}$ & 79.7$_{\smallsup{9.4}}$\\
\multirow{1}{*}{\fssc{XXL}} & \prompt, \fssc{FP} & 40.8$_{\smallsup{2.6}}$ & 90.0$_{\smallsup{3.0}}$ & 35.7$_{\smallsup{1.6}}$ & 2.2$_{\smallsup{1.5}}$ & 96.0$_{\smallsup{1.4}}$ & 29.0$_{\smallsup{0.5}}$ & 5.3$_{\smallsup{5.1}}$ & 85.3$_{\smallsup{5.1}}$ & 36.1$_{\smallsup{2.6}}$ & 0.6$_{\smallsup{0.5}}$ & 97.6$_{\smallsup{1.3}}$ & 36.9$_{\smallsup{1.2}}$ & 9.0$_{\smallsup{5.4}}$ & 80.8$_{\smallsup{9.3}}$ \\
\midrule
\multirow{1}{*}{\fssc{XXL}} & \model & 46.7$_{\smallsup{0.1}}$ & 94.4$_{\smallsup{0.8}}$ & 35.5$_{\smallsup{0.6}}$ & 9.1$_{\smallsup{3.1}}$ & 86.0$_{\smallsup{3.1}}$ & 27.2$_{\smallsup{0.4}}$ & 19.7$_{\smallsup{2.5}}$ & 57.5$_{\smallsup{2.8}}$ & 34.0$_{\smallsup{0.5}}$ & 14.8$_{\smallsup{3.5}}$ & 79.1$_{\smallsup{3.5}}$ & 30.1$_{\smallsup{0.5}}$ & 32.7$_{\smallsup{6.6}}$ & 16.8$_{\smallsup{3.6}}$\\
\multirow{1}{*}{\fssc{XXL}} & \model, \fssc{Mix-Unsup} &  46.7$_{\smallsup{0.1}}$ & 95.5$_{\smallsup{1.3}}$ & \textbf{39.5}$_{\smallsup{0.1}}$ & 2.2$_{\smallsup{0.4}}$ & 95.3$_{\smallsup{0.9}}$ & \textbf{32.3}$_{\smallsup{0.3}}$ & 6.2$_{\smallsup{1.1}}$ & 78.7$_{\smallsup{2.7}}$ & \textbf{38.3}$_{\smallsup{0.2}}$ & 1.5$_{\smallsup{0.7}}$ & 96.2$_{\smallsup{0.9}}$ & \textbf{32.4}$_{\smallsup{0.7}}$ & 17.0$_{\smallsup{2.1}}$ & 32.4$_{\smallsup{8.4}}$ \\
\multirow{1}{*}{\fssc{XXL}} & \model, \fssc{Mix-Unsup-All} & 46.3$_{\smallsup{0.1}}$ & 94.5$_{\smallsup{0.3}}$ & 38.2$_{\smallsup{0.1}}$ & 2.4$_{\smallsup{0.5}}$ & 95.2$_{\smallsup{0.7}}$ & 29.7$_{\smallsup{0.2}}$ & 13.0$_{\smallsup{0.3}}$ & 73.0$_{\smallsup{0.7}}$ & 37.8$_{\smallsup{0.2}}$ & 2.5$_{\smallsup{0.9}}$ & 93.4$_{\smallsup{0.5}}$ & 31.8$_{\smallsup{0.6}}$ & 17.4$_{\smallsup{1.6}}$ & 45.2$_{\smallsup{4.0}}$\\
\multirow{1}{*}{\fssc{XXL}} & \model, \fssc{IT-Gigaword} & 46.3$_{\smallsup{0.2}}$ & 95.6$_{\smallsup{0.4}}$ & 24.8$_{\smallsup{0.2}}$ & 81.2$_{\smallsup{1.9}}$ & 9.9$_{\smallsup{1.4}}$ & 20.8$_{\smallsup{0.1}}$ & 78.8$_{\smallsup{1.9}}$ & 3.8$_{\smallsup{0.2}}$ & 31.3$_{\smallsup{0.3}}$ & 32.5$_{\smallsup{4.1}}$ & 54.4$_{\smallsup{4.1}}$ & 22.8$_{\smallsup{0.2}}$ & 87.2$_{\smallsup{0.9}}$ & 2.4$_{\smallsup{0.6}}$\\
\multirow{1}{*}{\fssc{XXL}} & \model, \fssc{IT-LM} & 46.3$_{\smallsup{0.0}}$ & 95.2$_{\smallsup{0.9}}$ & 25.7$_{\smallsup{0.1}}$ & 72.7$_{\smallsup{5.2}}$ & 16.6$_{\smallsup{4.5}}$ & 22.4$_{\smallsup{0.2}}$ & 59.5$_{\smallsup{1.6}}$ & 15.7$_{\smallsup{1.3}}$ & 19.5$_{\smallsup{4.1}}$ & 82.0$_{\smallsup{14.2}}$ & 10.8$_{\smallsup{12.5}}$ & 25.1$_{\smallsup{0.1}}$ & 66.5$_{\smallsup{1.1}}$ & 6.4$_{\smallsup{1.4}}$ \\
\bottomrule
\end{tabular}
\end{adjustbox}
\caption{Summarization quality (\ssc{SP-Rouge}) and language identification confidence scores (\ssc{LID}) across two model sizes (\ssc{Base} and \ssc{XXL}) and methods (numbers in the subscript indicate the standard deviation across 3 random seeds). Mixing in unlabeled multilingual data (\ssc{Mix-Unsup}/\ssc{Mix-Unsup-All}) helps prevent catastrophic forgetting for \modeltuning. Intermediate tuning (\mbox{\ssc{IT-Gigaword}}/\mbox{\ssc{IT-LM}}) does not result in reliable gains. Factorized prompts (\mbox{\ssc{FP-En}}/ \mbox{\ssc{FP}}) lead to an improvement in target language accuracy, and an improvement in \ssc{SP-Rouge} in cases where vanilla \prompttuning shows the worst performance. 
}
\label{tbl:methods}
\end{table*}
Table~\ref{tbl:methods} shows our experiment results for different approaches described in~\S\ref{section:method_description}. As can been seen, mixing in unlabeled multilingual data (\ssc{Mix-Unsup}/\ssc{Mix-Unsup-All}) helps prevent catastrophic forgetting for \modeltuning. Intermediate tuning (\mbox{\ssc{IT-Gigaword}}/\mbox{\ssc{IT-LM}}) does not result in reliable gains. Finally, factorized prompts (\mbox{\ssc{FP-En}}/ \mbox{\ssc{FP}}) lead to an improvement in target language accuracy, and an improvement in \sprouge in cases where vanilla \prompttuning shows the worst performance.  

\section{Intermediate tuning}
\label{appendix:intermediate_tuning}
As an adaptation step, we perform model or prompt tuning on an intermediate task before training on \ssc{Wikilingua-0}. Intermediate tuning has been used to boost performance on English tasks for both \modeltuning~\cite{JPhang19,TVu20} and \prompttuning~\cite{TVu22}, and has been successfully applied to the zero-shot cross-lingual transfer setting~\cite{JPhang20,KMaurya21} for \modeltuning. \citet{KMaurya21} show that intermediate tuning on an auxiliary unsupervised task from the target language is helpful in conjunction with freezing some model components for \ssc{ModelTuning}. Previous work has used an auxiliary task designed to be close to the main task, while we simply use
\mcfour data. For each target language we create a causal, left-to-right LM task by providing no context, i.e., the encoder's input is empty (\bssc{IT-LM}). %
To further explore the effect of continued training on English data, we include an additional experiment where the \ssc{Gigaword}~\cite{DGraff03} summarization dataset is used as the intermediate task (\bssc{IT-Gigaword}).\footnote{We found that additional tuning was helpful for intermediate tuning on large datasets. As such, we performed $200{,}000$ steps during tuning on an intermediate task and selected the best prompt checkpoint based on validation performance on that task.}

\paragraph{Intermediate tuning does not give reliable gains:} As can be seen in Table~\ref{tbl:methods}, intermediate tuning on English summarization (\mbox{\ssc{IT-Gigaword}}) improves English performance, but generally hurts \ssc{XGen} capabilities. For \modeltuning, it exacerbates catastrophic forgetting and harms overall performance across all model sizes. For \prompttuning, English intermediate tuning provides small gains at \ssc{Base} size, but is harmful at \ssc{XXL} size. Intermediate tuning on an \ssc{LM} task in the target language (\mbox{\ssc{IT-LM}}) has a neutral or negative effect in most cases, running somewhat counter to the findings of \citet{KMaurya21}.\footnote{Note, however that their unsupervised task was designed to be well-aligned with their downstream tasks of choice.} Compared to directly mixing in unlabeled multilingual data, intermediate tuning has little benefit on language accuracy. This smaller effect is to be expected, given that the final stage of English-only training is still ample opportunity to overfit on English and catastrophically forget other languages.

\end{document}